\documentclass[runningheads]{llncs}

\usepackage{eccv}

\usepackage{eccvabbrv}

\usepackage{graphicx}
\usepackage{booktabs}

\usepackage{multirow}
\usepackage[table,xcdraw]{xcolor}
\usepackage{colortbl}
\newcommand{\gc}{\cellcolor[HTML]{F5F5F5}}

\usepackage{hyperref}

\usepackage{marvosym}

\begin{document}

\title{DualDiff3D: Dual Structure-Appearance Diffusion Priors for Reliability-Enhanced 3D Gaussian Splatting} 

\titlerunning{DualDiff3D}

\author{Qian Wang\inst{1,2,\text{\textasteriskcentered}}
\and
Yu Wang\inst{1,\text{\textasteriskcentered}} 
\and
Weiqi Li\inst{1,2}
\and
Xinhua Cheng\inst{1}
\and
Xiandong Meng\inst{2}
\and
Ronggang Wang\inst{1,2,3}
\and
Jian Zhang\inst{1,2,3,\text{\Letter}}
}

\authorrunning{Q.~Wang et al.}

\institute{School of Electronic and Computer Engineering, Peking University, Shenzhen, China 
\and
Pengcheng Laboratory, Shenzhen, China
\and
Guangdong Provincial Key Laboratory of Ultra High Definition Immersive Media Technology, Shenzhen, China \\
\url{https://akaneqwq.github.io/dualdiff3d}
}

\maketitle

\let\thefootnote\relax\footnotetext{
\text{\textasteriskcentered} Equal contribution \\
\text{\Letter} Corresponding author: zhangjian.sz@pku.edu.cn 
}

{
\centering
\includegraphics[width=\textwidth]{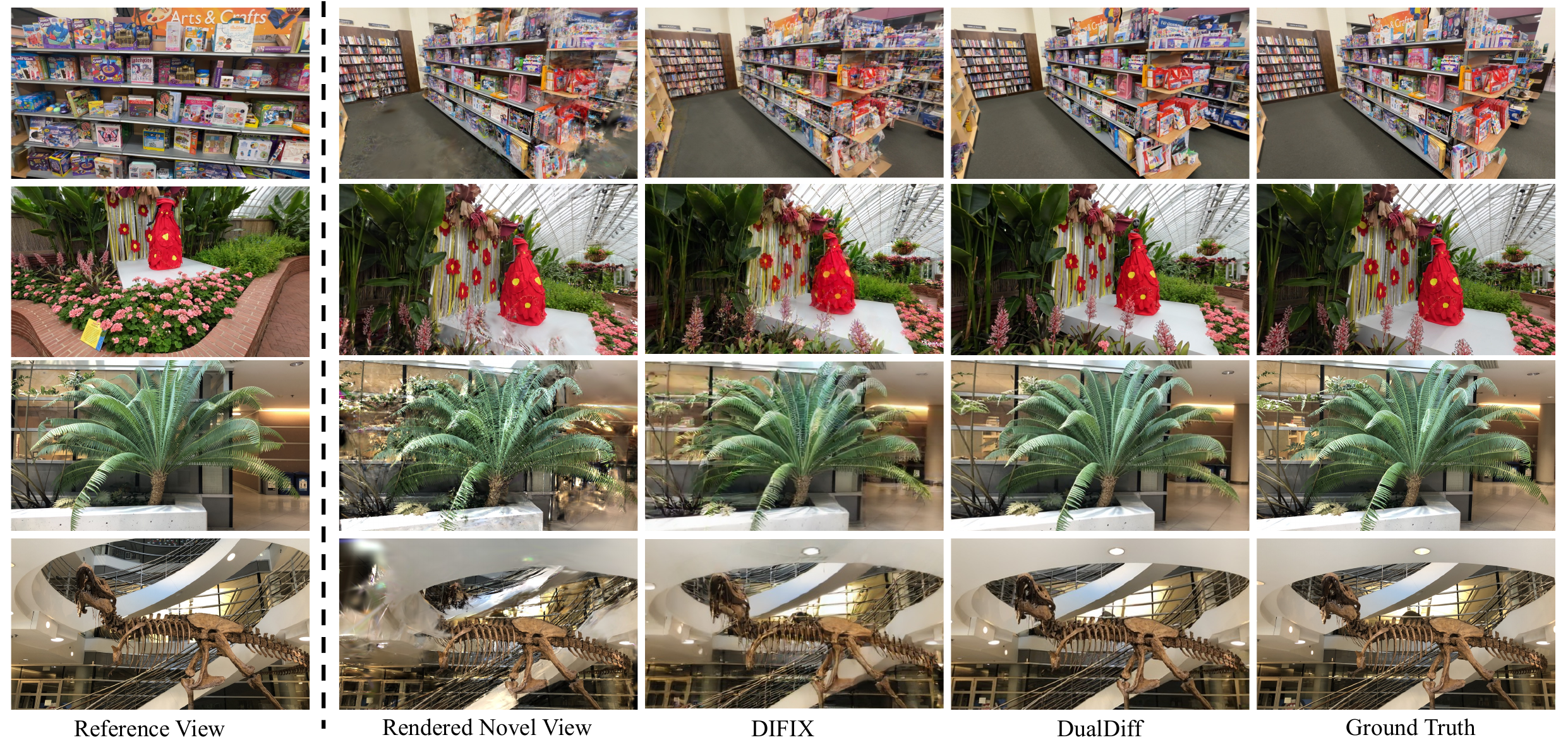}
\captionsetup{type=figure}
\caption{\textbf{Comparison of refined results for novel views with artifacts.} DualDiff maintains structural consistency with the rendered novel view while matching the appearance of the leftmost reference view, achieving better performance.}
\label{fig:teaser}
}
\begin{abstract}
While 3D Gaussian Splatting (3DGS) has revolutionized 3D reconstruction and novel-view synthesis, scenarios with limited input views often lead to poor reconstruction quality and artifacts in rendered novel views. 
Recent efforts attempt to utilize powerful diffusion priors, yet they typically process rendered and reference views concatenated along an additional dimension in a single network.
These methods overlook an inherent nature that different views should maintain appearance similarity but differ in structure due to view shifts, leading to blur caused by conflicts between the two properties.
In this paper, we propose \textbf{DualDiff}, a novel pipeline that leverages dual diffusion priors with a Structure-Appearance Attention (SAA) module to introduce reference guidance for refining low-quality novel views rendered from flawed 3D representations. 
Specifically, we retain one diffusion branch to focus on extracting structural information from the low-quality novel views, while introducing another branch to ensure appearance consistency with reference views.
Furthermore, we present a 3D reconstruction framework named \textbf{DualDiff3D}, which integrates a reliability-enhanced Render-Refine-Optimize (RRO) loop to progressively and robustly incorporate the refined novel views, yielding more accurate 3DGS. 
Extensive experiments demonstrate that our approach outperforms state-of-the-art methods even in the inference-only setting, with further performance gains achievable through training. 
Our code and pre-trained weights are available at \url{https://github.com/Akaneqwq/DualDiff3D}.
\end{abstract}    
\section{Introduction}
\label{sec:intro}

3D reconstruction and novel-view synthesis (NVS)~\cite{nerf,nerfactor,mipnerf,zipnerf,tensorf} are important tasks in computer vision that aim to generate realistic images of a scene from previously unseen viewpoints by giving a limited set of input images, which drives diverse applications such as virtual/augmented reality (VR/AR), autonomous driving, and digital content creation.
In recent years, breakthroughs like 3D Gaussian Splatting (3DGS)~\cite{3dgs,fsgs} have demonstrated remarkable capability to produce photorealistic renderings.
However, methods developed based on 3DGS encounter challenges in scenarios with sparse inputs for synthesizing novel views that are far from the input views, due to insufficient information and constraints provided by sparse viewpoints.

To address the limitations in scenes with sparse views, recent works~\cite{3dgsenhancer,difix3d} have explored generative models, particularly diffusion models, for enhancing 3D reconstruction and NVS.
These methods typically construct an initial low-quality 3D representation from sparse views, render a series of novel views, and then take the original sparse views as references to refine these rendered novel views.
The refined novel views are subsequently incorporated into the training set for subsequent 3DGS optimization.
Based on the type of diffusion prior adopted by the refinement model, these approaches can be categorized into two classes.

One class, exemplified by 3DGS-Enhancer~\cite{3dgsenhancer}, leverages video diffusion priors for refinement. 
They sample sequentially along a trajectory containing both reference views and novel views, then concatenate sampled views along the temporal dimension and feed them into the video diffusion model to achieve enhanced results. 
These works leverage the inherent temporal continuity of video models to introduce cross-view consistency constraints.
However, high inference latency and computational cost make them impractical for online 3DGS optimization.

\begin{figure}[t]
    \centering
    \captionsetup{type=figure}
    \includegraphics[width=0.9\linewidth]{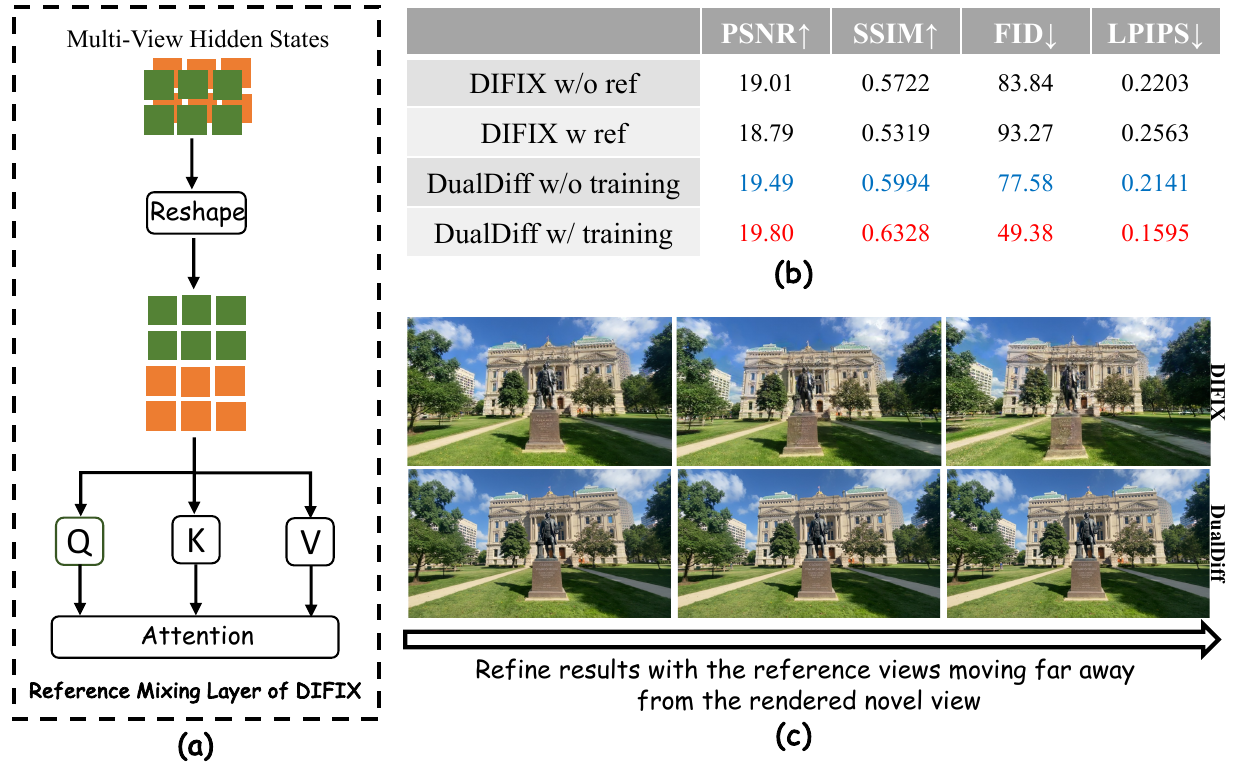}
    \captionsetup{type=figure}
    \caption{\textbf{Motivation of our method.} (a) Illustrates the reference fusion adopted by DIFIX. (b) Shows that on the LLFF dataset, DIFIX achieves better performance in reference-free generation than in reference-based generation; in contrast, DualDiff yields performance gains merely through structural modification without any additional training. (c) Presents the subjective results of DIFIX and DualDiff as the reference view becomes increasingly distant, where DualDiff can effectively avoid blurring.}
    \label{fig:motivation}
\end{figure}

The other class, such as DIFIX3D+~\cite{difix3d}, adopts image diffusion priors for refinement, which reduces both inference latency and computational overhead. 
To enable 2D image diffusion models to handle the additional view dimension, these methods directly embed the view dimension into the batch dimension. 
For supporting cross-view reference, they concatenate features from multiple views along the spatial dimension before the self-attention module, and restore the view dimension to the batch dimension after this module, as shown in \cref{fig:motivation}(a).
Although such a method is a straightforward strategy for introducing cross-view information interaction into image diffusion models, we identify notable drawbacks in this design, \ie, novel views and reference views provide distinct information in 3D reconstruction and NVS tasks. 
Given a rendered novel view with artifacts, we hope the refined result to maintain consistency with the original novel view in terms of perspective (referred to as structural consistency), while filling the artifact regions with appropriately warped content from corresponding parts of the reference views (referred to as appearance consistency). 
Thus, we clarify that handling both structural and appearance information using a single network is inappropriate. 
As illustrated in \cref{fig:motivation}(b), the evaluation results of DIFIX3D+ verify our argument. 
Surprisingly, the refinement performance with reference views is inferior to that without references. 
Furthermore, as the angular distance between reference views and novel views increases, the blurriness caused by conflicts between structural and appearance information in refined results becomes increasingly pronounced.

Therefore, we propose DualDiff, a novel pipeline that addresses these limitations by leveraging dual diffusion priors with a Structure-Appearance Attention (SAA) module to introduce reference guidance as presented in \cref{fig:dualdiff}. 
Specifically, we retain one diffusion branch to focus on extracting structural information from the low-quality rendered novel views, and introduce another diffusion branch to ensure appearance consistency with reference views. 
During refinement, we concatenate the hidden states from the self-attention module of the appearance branch to those from structure branch to produce new key and value, transforming the self-attention module of structure branch into SAA module to enable information interaction between dual branches.
We also interestingly discover that the proposed DualDiff pipeline can be regarded as a ``free lunch'' for existing reference-based novel view refinement models leveraging diffusion priors. 
When using the DualDiff pipeline, even if the weights of the U-Nets in both branches are only initialized with those from DIFIX3D+ without any additional training, it can still yield a $0.7\,\mathrm{dB}$ gain in PSNR, as shown in \cref{fig:motivation}(b). 
This validates the rationality of our approach, decoupling structural and appearance processing while fusing information via attention mechanisms.

Furthermore, we present a 3D reconstruction framework named DualDiff3D, which integrates a reliability-enhanced Render-Refine-Optimize (RRO) loop equipped with a Progressive Sampling and Filtering (PSF) strategy and a Confidence-Driven Weighting (CDW) process. This cyclic process iteratively strengthens the 3D representation by incorporating increasingly high-quality novel view constraints.

Finally, our contributions can be summarized as follows: 
\begin{itemize}
\item We propose DualDiff, a novel two-branch pipeline to refine low-quality rendered novel views, guided by reference views. We are the first to recognize the necessity of decoupling structure and appearance for this task. By fully leveraging dual diffusion priors and introducing a Structure-Appearance Attention (SAA) module, our model independently ensures structural consistency and appearance consistency with distinct inputs.
\item We introduce DualDiff3D, a 3D reconstruction framework based on a reliability-enhanced Render-Refine-Optimize (RRO) loop, integrated with a Progressive Sampling and Filtering (PSF) strategy and a Confidence-Driven Weighting (CDW) process. We effectively address the critical instability issue caused by directly feeding refined outputs from diffusion models into 3DGS optimization, achieving a logical and complete closed-loop solution.
\item Extensive experiments on benchmark datasets demonstrate that DualDiff3D improves the quality of 3D reconstruction and novel view synthesis (NVS) in both objective metrics and subjective quality even in the inference-only scenario, with further performance gains attainable via fine-tuning.
\end{itemize}

\section{Related Works}
\label{sec:related}

\subsection{Diffusion Models}
The Denoising Diffusion Probabilistic Model~\cite{ddpm,ddim} has proven to be highly successful in generating high-quality images, outperforming previous approaches such as generative adversarial networks (GANs)~\cite{gan, stackgan}, variational autoencoders (VAEs)~\cite{vae, vqvae}, and flow-based methods~\cite{flow}. 
With text guidance during training, users can generate images based on textual input. Noteworthy examples include GLIDE~\cite{glide}, which utilizes textual prompts as conditions and adopts classifier-free guidance~\cite{cfg}. 
DALLE-2~\cite{dalle2} leverages the latent space of CLIP~\cite{clip} as a condition, and Imagen~\cite{imagen} injects features from a large language model into cascaded diffusion models. 
To address the computational burden of the iterative denoising process, LDM~\cite{ldm} conducts the diffusion process on a compressed latent space rather than the original pixel space, achieving success in reducing computational requirements. 
This accomplishment has prompted further exploration in customization~\cite{dreambooth, textinversion}, image guidance~\cite{paint}, precise control~\cite{controlnet, t2iadapter}, video generation~\cite{wang2024360dvd,li2026omnidrag,yang2026gencompositor}, and image restoration and quality assessment~\cite{li2026uare,li2026q}.

\subsection{3D Reconstruction and NVS Methods}
3D reconstruction and NVS have evolved around balancing rendering fidelity, efficiency, and scalability.
Traditional approaches relied on explicit primitives or early implicit functions~\cite{deepsdf, occupancy}, which either suffered from high memory costs or failed to jointly model geometry and appearance, limiting realism and scalability. 
Typical works like KinectFusion~\cite{KinectFusion} realized real-time dense surface mapping but were confined to small-scale scenes.
Neural Radiance Fields (NeRF)~\cite{nerf, mipnerf, zipnerf, tensorf, plenoxels, instantngp, cheng2023panoptic} revolutionized the field by encoding scene geometry and appearance into a continuous implicit neural network. 
Despite generating photorealistic novel views, NeRF and its variants faced critical bottlenecks in inference speed and computational efficiency, hindering real-time applications.
3DGS~\cite{3dgs} later emerged as a landmark explicit method that combines the advantages of implicit neural modeling and traditional point-based representations. It enables real-time photorealistic rendering but struggles with novel views far from input perspectives under sparse-view conditions.
Recent studies have also investigated the security of 3DGS representations~\cite{zhang2024gs,zhang2025securegs}.

\subsection{Diffusion Priors for NVS}
Diffusion priors have emerged as a transformative force in novel view synthesis, offering robust solutions to challenges like sparse input views and geometric inconsistency. 
A key application lies in enhancing 3D representations with view consistency. 3DGS-Enhancer~\cite{3dgsenhancer} integrates 2D video diffusion priors into 3DGS, reformulating view consistency as temporal coherence in video generation. 
Similarly, GaussianSR~\cite{gaussiansr} employs 2D diffusion priors via score distillation sampling to super-resolve low-resolution inputs, addressing redundancy in 3D Gaussian primitives through timestep range shrinking and primitive pruning.
Diffusion priors have also been adopted for image-to-3D generation, including ~\cite{yang2025hybrid,cheng2023progressive3d,zhang2024repaint123}.
For dynamic scenes, DiffusionPriors~\cite{diffusionprior} fine-tunes an RGB-D diffusion model on monocular video frames, distilling knowledge into 4D neural radiance fields to disentangle motion and structure—even with unknown camera poses. 
~\cite{yang20264dvd, cheng2026360explorer} similarly leverage video diffusion for high-quality 4D content generation.
Handling extreme sparsity, DreamSparse~\cite{dreamsparse} combines a geometry module (capturing 3D spatial priors) with a spatial guidance model to convert rendered features, guiding 2D diffusion models toward geometrically consistent novel views. 
DIFIX3D+~\cite{difix3d} trains a single-step diffusion model to remove artifacts in rendered novel views caused by suboptimal 3D representation.
\section{Preliminary}
\label{sec:pre}

\subsection{3D Gaussian Splatting}
3DGS is an innovative and state-of-the-art approach in the field of 3D reconstruction and NVS. 
Distinguished from implicit representation methods such as NeRF~\cite{nerf}, which utilize volume rendering, 3DGS leverages the splatting technique~\cite{yifan2019differentiable} to generate images, achieving remarkable real-time rendering speed. 
Specifically, 3DGS represents the scene through a set of anisotropic Gaussians, defined with its center position $\boldsymbol{\mu} \in \mathbb{R}^3$, covariance $\mathbf{\Sigma} \in \mathbb{R}^{3\times3}$ which can be decomposed into scaling factor $\boldsymbol{s} \in \mathbb{R}^3$ and rotation factor $\boldsymbol{q} \in \mathbb{R}^4$, color defined by spherical harmonic coefficients $\boldsymbol{h} \in \mathbb{R}^{3\times (k+1)^2}$ (where $k$ represents the order of spherical harmonics), and opacity $\alpha \in \mathbb{R}^1$. Then, the 3D Gaussian can be queried as follows:
\begin{equation}
	\label{eq: gaussian}
	\mathcal{G}(\mathbf{x})=\operatorname{e}^{-\frac{1}{2}(\mathbf{x}-\boldsymbol{\mu})^{\top}\mathbf{\Sigma}^{-1}(\mathbf{x}-\boldsymbol{\mu})},
\end{equation}
where $\mathbf{x}$ represents the position of the query point. Subsequently, an efficient 3D to 2D Gaussian mapping~\cite{zwicker2002ewa} is employed to project the Gaussian onto the image plane:
\begin{equation}
	\hat{\boldsymbol{\mu}}={\mathbf{P}\mathbf{W}\boldsymbol{\mu}}, \quad \hat{\mathbf{{\Sigma}}}=\mathbf{J}\mathbf{W} \mathbf{\Sigma} \mathbf{W^{\top}}\mathbf{J^{\top}},
\end{equation}
where $\hat{\boldsymbol{\mu}}$ and $\hat{\mathbf{\Sigma}}$ separately represent the 2D mean position and covariance of the projected 3D Gaussian. 
$\mathbf{P}$, $\mathbf{W}$ and $\mathbf{J}$ denote the projective transformation, viewing transformation, and Jacobian of the affine approximation of $\mathbf{P}$, respectively. 
The color of the pixel on the image plane, denoted by $\mathbf{p}=(u, v)$, uses a typical neural point-based rendering~\cite{kopanas2022neural}. Let $\mathbf{C} \in \mathbb{R}^{H\times W\times3}$ represent the color of the rendered image, where $H$ and $W$ represent the height and width of the image. The rendering process is outlined as follows:
\begin{equation}
\begin{split}
	{\mathbf{C}[\mathbf{p}]} &= {\sum_{i=1}^{N}
	\boldsymbol{c}_{i}\sigma_{i}
	\prod_{j=1}^{i-1}(1-\sigma_{j})},
     \quad \\
    \sigma_i &= \alpha_i\operatorname{e}^{-\frac12(\mathbf{p}-\hat{\boldsymbol{\mu}})^{\top}\hat{\mathbf{\Sigma}}^{-1}(\mathbf{p}-\hat{\boldsymbol{\mu}})},
\label{eq_render}
\end{split}
\end{equation}
where $N$ represents the number of Gaussians that overlap the pixel $\mathbf{p}$. 
$\boldsymbol{c}_{i} \in \mathbb{R}^{3}$ and ${\alpha}_{i} \in \mathbb{R}^1$ denote the color calculated from $\boldsymbol{h}_i$ and opacity of the $i$-th Gaussian.

\subsection{Diffusion Model}
Given an input signal $x_{0}$, a diffusion forward process in DDPM~\cite{ddpm} is defined as:
\begin{equation}
    q(x_t | x_{t-1}) = \mathcal{N}(x_t; \sqrt{1 - \beta_{t}}x_{t-1}, \beta_{t}I),
\end{equation}
for $t = 1, ..., T$, where $T$ is the total timestep of the diffusion process. 
A noise depending on the variance $\beta_{t}$ is gradually added to $x_{t-1}$ to obtain $x_t$ at the next timestep and finally reach $x_T \sim \mathcal{N}(0, I)$. 
The goal of the diffusion model is to learn to reverse the diffusion process (denoising). 
Given a random noise $x_t$, the model predicts the added noise at the next timestep $x_{t-1}$ until the origin signal $x_0$:
\begin{equation}
    p_\theta(x_{t-1} | x_t) = \mathcal{N}(x_{t-1}; \mu_\theta(x_t, t), \Sigma_\theta(x_t, t)),
\end{equation}
for $t = T, ..., 1$. 
We fix the variance $\Sigma_\theta(x_t, t)$ and utilize the diffusion model with parameter $\theta$ to predict the mean of the inverse process $\mu_\theta(x_t, t)$. 
The model can be simplified as denoising models $\epsilon_\theta(x_t, t)$, which are trained to predict the noise of $x_t$ with a noise prediction loss:
\begin{equation}
    \mathcal{L} = \mathbb{E}_{x_0, y, \epsilon \sim \mathcal{N}(0, I), t}[\|\epsilon - \epsilon_\theta(x_t, t, \tau_\theta(y))\|^2_2],
\end{equation}
where $\epsilon$ is the added noise to the input image $x_0$, $y$ is the corresponding textual description, $\tau_\theta(\cdot)$ is a text encoder mapping the string to a sequence of vectors.

\section{DualDiff}
\label{sec:dualdiff}

\begin{figure*}[t]
    \centering
    \captionsetup{type=figure}
    \includegraphics[width=1.0\linewidth]{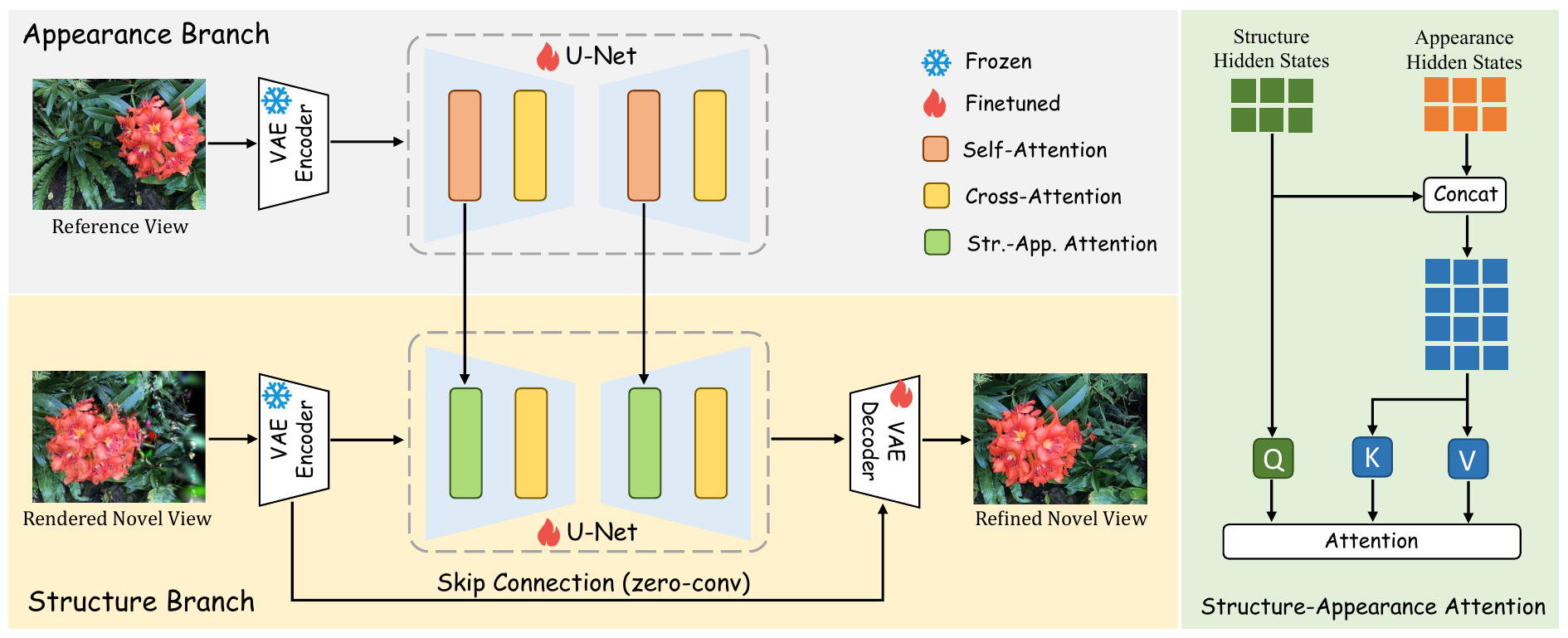}
    \captionsetup{type=figure}
    \caption{\textbf{Pipeline of DualDiff.} DualDiff leverages dual diffusion branches integrated with a Structure-Appearance Attention (SAA) module to incorporate reference guidance, thereby refining low-quality novel views rendered from defective 3D representations.}
    \label{fig:dualdiff}
\end{figure*}

\subsection{Inspiration}
The design of DualDiff mainly draws inspiration from two sources. 
First, we are inspired by the latest dual diffusion branch design in other image-to-image generation tasks such as virtual try-on and image animation~\cite{animateanyone}.
For example, virtual try-on tasks commonly adopt a dual-branch architecture to separately ensure structural alignment with the model image and appearance alignment with the garment image, and this design logic closely parallels our novel view refinement task. 
The success in these fields demonstrates the unique advantages of dual-branch architectures in addressing tasks requiring the decoupling of appearance and structural information. 
Furthermore, it also validates that diffusion models trained on large datasets are naturally effective feature extractors, which can be fine-tuned to focus on specific types of image characteristics.
Second, we are inspired by analyzing two primary condition injection frameworks in diffusion models. 
Methods like ControlNet~\cite{controlnet} and T2I-Adapter~\cite{t2iadapter} use feature addition to merge latent and conditional representations, excelling at pixel-aligned tasks (\eg, skeleton-guided generation) due to element-wise fusion at consistent spatial locations. 
In contrast, methods like IP-Adapter~\cite{ipadapter} inject conditional features into attention mechanisms, enabling semantic region focus that suits appearance-aligned tasks (\eg, style transfer). Leveraging this distinction, we adopt attention-based injection for appearance-aligned reference views.

\subsection{Overview}
An overview of the DualDiff pipeline is presented in \cref{fig:dualdiff}, which is composed of a structure branch and an appearance branch. 
The denoising U-Net in both branches adopts a structure identical to the single-step diffusion model SD-Turbo~\cite{sdturbo} and weights initialized with DIFIX~\cite{difix3d}. 
Following DIFIX, our model takes rendered novel views as input instead of Gaussian noise and sets a lower timestep $t = 200$, corresponding to a lower noise level. This design allows the refinement process to simulate a one-step denoising procedure under low-noise conditions.
We utilize the same frozen VAE encoder to encode both rendered novel views and reference views into the latent space, which are then fed into the U-Net. 
Only the structure branch is equipped with a VAE decoder to decode the final refined novel views from the latent space to pixel space.
To mitigate information loss caused by VAE encoding and decoding, we add zero-initialized skip connections between corresponding layers of the VAE encoder and decoder in the structure branch, and fine-tune the VAE decoder with LoRA~\cite{lora}.

\subsection{Structure-Appearance Attention Module}
To incorporate the appearance guidance from reference views into the structure branch, we replace the self-attention module in the denoising U-Net with our proposed Structure-Appearance Attention (SAA) module. 
As illustrated in \cref{fig:dualdiff}, given the hidden states $\mathbf{H}_\text{app} \in \mathbb{R}^{n \times d}$ from the self-attention module of one U-Net block in the appearance branch, and the corresponding hidden states $\mathbf{H}_\text{str} \in \mathbb{R}^{m \times d}$ from the structure branch, we concatenate them along the sequence length dimension to obtain mixed hidden states that fuse structural and appearance features, denoted as $\mathbf{H}_\text{mix} \in \mathbb{R}^{(m+n) \times d}$.
Since our goal is to inject information from the reference view into the novel view, we set the queries $\mathbf{Q} \in \mathbb{R}^{m \times d}$ to be derived from $\mathbf{H}_\text{str}$, while the keys $\mathbf{K} \in \mathbb{R}^{(m+n) \times d}$ and values $\mathbf{V} \in \mathbb{R}^{(m+n) \times d}$ are derived from $\mathbf{H}_\text{mix}$.
This process can be formulated as:
\begin{equation}
\begin{split}
\mathbf{Q} &= \mathbf{W}_q \mathbf{H}_\text{str}, \\
\mathbf{K} &= \mathbf{W}_k \mathbf{H}_\text{mix}, \\
\mathbf{V} &= \mathbf{W}_v \mathbf{H}_\text{mix},
\end{split}
\end{equation}
where $\mathbf{W}_q$, $\mathbf{W}_k$ and $\mathbf{W}_v$ are learnable projection matrices.
The structure-appearance attention is finally calculated as:
\begin{equation}
\text{Attn}(\mathbf{Q}, \mathbf{K}, \mathbf{V}) = \text{Softmax}\left(\frac{\mathbf{Q}\mathbf{K}^{\mathrm{T}}}{\sqrt{d}}\right) \mathbf{V}.
\end{equation}

\subsection{Loss Function}
Although our pipeline can achieve improvements in a training-free manner when initialized with the pre-trained weights of DIFIX, it can still benefit from training. 
We adopt the same loss function as DIFIX, which consists of an L2 reconstruction loss between the refined novel views and the ground truth, a perceptual LPIPS loss for better visual quality, and a Gram-Matrix loss based on features of VGG-16~\cite{vgg} that encourages sharper details:
\begin{equation}
\mathcal{L} = \mathcal{L}_\text{Recon} + \mathcal{L}_\text{LPIPS} + \lambda_{0}\mathcal{L}_\text{Gram},
\end{equation}
where $\lambda_{0}$ is a hyper-parameter set to $0.1$ to weight Gram-Matrix loss.
\section{DualDiff3D}
\label{sec:dualdiff3d}

\subsection{Reliability-Enhanced RRO Loop}
In scenarios with limited available views, 3DGS often yields degraded reconstructions due to insufficient scene coverage and high sensitivity to geometric inconsistencies. 
To address this, we propose the Render-Refine-Optimize (RRO) loop, which leverages our DualDiff pipeline to augment 3DGS’s training set with refined novel views—generated from renderings of the initial low-quality 3DGS scene representation. 
The RRO loop consists of three core steps: 
(1) Render Step: render batches of novel views from the currently reconstructed 3DGS; 
(2) Refine Step: utilize the DualDiff pipeline to refine details and eliminate artifacts in sampled novel views; 
(3) Optimize Step: employ the refined novel views to further optimize the 3DGS model. 
However, direct incorporation of these refined images introduces three fundamental challenges: 
(1) Refined images inevitably exhibit minor geometric inconsistencies with ground-truth observations, which can destabilize 3DGS optimization; 
(2) 3DGS is highly sensitive to such geometric deviations, particularly when novel views are far from the training distribution; 
(3) Novel view selection is non-trivial, as arbitrary sampling may provide limited information gain while amplifying reconstruction errors. 
To build a more reliable and robust RRO loop, we integrate a Progressive Sampling and Filtering (PSF) strategy and a Confidence-Driven Weighting (CDW) process. 

\begin{figure*}[t]
    \centering
    \captionsetup{type=figure}
    \includegraphics[width=1.0\linewidth]{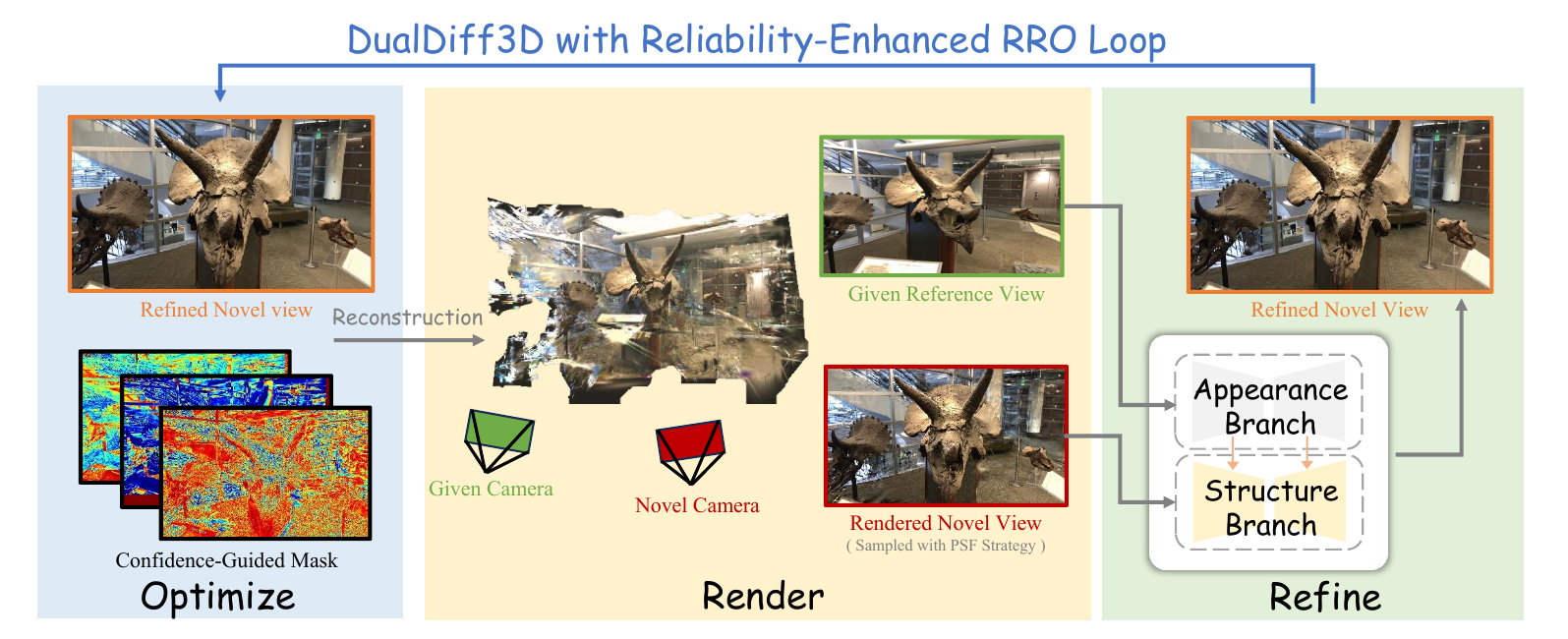}
    \captionsetup{type=figure}
    \caption{\textbf{Pipeline of DualDiff3D.} DualDiff3D integrates a reliability-enhanced RRO loop to progressively and robustly incorporate the refined novel views, yielding more accurate 3DGS.}
    \label{fig:dualdiff3d}
\end{figure*}

\subsection{Progressive Sampling and Filtering Strategy}
The PSF strategy progressively expands the training set while preserving geometric consistency. Instead of randomly sampling novel viewpoints, PSF interpolates and slightly extrapolates between existing training cameras, introducing small perturbations to enhance diversity. Each sampled novel view is evaluated using a composite confidence mask (defined in \cref{sec:cdw}) at both pixel and image levels. Only high-confidence views exhibiting strong geometric and appearance consistency are admitted into the augmented training pool.
The renderings of these sampled novel views, paired with their closest ground-truth observations, are then fed into the DualDiff pipeline for refinement.

\subsection{Confidence-Driven Weighting Process}
\label{sec:cdw}
The CDW process aims to evaluate refined novel views via a composite confidence-guided mask, ensuring only high-confidence regions are incorporated into the 3DGS training process. 
The confidence-guided mask $M \in [0,1]^{H \times W}$ integrates three complementary confidence measures:

\paragraph{Refine Variance Confidence.} 
Each novel view is refined $K$ times by DualDiff refinement model, producing $\{I^{(k)}_{\text{refine}}\}_{k=1}^K$. 
The per-pixel variance is used to estimate uncertainty:
\begin{equation}
\begin{split}
\sigma^2(u) &= \frac{1}{K}\sum_{k=1}^{K} \big(I^{(k)}_{\text{refine}}(u) - \bar{I}_{\text{rep}}(u)\big)^2, \\
\quad 
\bar{I}_{\text{rep}}(u) &= \frac{1}{K}\sum_{k=1}^{K} I^{(k)}_{\text{refine}}(u),
\end{split}
\end{equation}
where $u$ denotes a pixel coordinate. 
The corresponding confidence is defined as: 
\begin{equation}
C_{\text{var}}(u) = \exp(-\alpha \cdot \sigma^2(u)),
\end{equation}
where $\alpha$ controls sensitivity to variance.

\paragraph{Pixel Discrepancy Confidence.} 
Given the original low-quality rendering $I_{\text{low}}$, the absolute difference with the refined result provides another confidence cue:
\begin{equation}
\begin{split}
d(u) &= \| I_{\text{refine}}(u) - I_{\text{low}}(u) \|_1, \quad \\
C_{\text{diff}}(u) &= \exp(-\beta \cdot d(u)),
\end{split}
\end{equation}
where $\beta$ scales the influence of discrepancies.

\paragraph{Reprojection Confidence.} 
Using the rendered depth map $D_{\text{novel}}$, pixels in the refined novel views are reprojected into overlapping reference images $\{I_{\text{ref}}^j\}$. For each valid projection, the reprojection error is:
\begin{equation}
e_{\text{proj}}(u) = \frac{1}{N_u}\sum_{j=1}^{N_u} 
\big\| I_{\text{refine}}(u) - I_{\text{ref}}^j(\pi_j(u, D_{\text{novel}}(u))) \big\|_1,
\end{equation}
where $\pi_j(\cdot)$ is the projection operator into reference views $j$, and $N_u$ is the number of valid projections. 
The confidence is:
\begin{equation}
C_{\text{proj}}(u) = \exp(-\gamma \cdot e_{\text{proj}}(u)),
\end{equation}
with $\gamma$ controlling sensitivity.

\paragraph{Training Process.} 
The final composite confidence-guided mask is:
\begin{equation}
\begin{split}
M(u) = \lambda_1 &C_{\text{var}}(u) + \lambda_2 C_{\text{diff}}(u) + \lambda_3 C_{\text{proj}}(u), \quad \\
&\lambda_1 + \lambda_2 + \lambda_3 = 1.
\end{split}
\end{equation}
During optimization, the loss for refined views is defined as
\begin{equation}
\mathcal{L}_{\text{refine}} = 
\sum_{u} M(u) \cdot \| I_{\text{refine}}(u) - I_{\text{rend}}(u) \|_1
+ \eta \cdot \text{LPIPS}(I_{\text{refine}}, I_{\text{rend}}),
\end{equation}
where $I_{\text{rend}}$ is the rendered image from the current 3DGS model and $\eta$ balances pixel-wise and perceptual losses. 
To ensure more stable training, each training batch always includes a certain ratio of original training views alongside novel views. 
Additional validation-and-rollback mechanism monitors the model’s performance on a held-out validation set, discarding newly added refined views if the 3D reconstruction quality deteriorates.
\section{Experiments}
\label{sec:exp}

\subsection{Implementation Details}
\paragraph{DualDiff.}
DualDiff is trained on 90\% of randomly selected scenes from the DL3DV~\cite{dl3dv} benchmark dataset. 
For each scene, we conduct vanilla 3DGS with 3, 6, 9, and 24 views, respectively, for 30k iterations. 
Rendered novel views from the reconstructions, together with their corresponding ground truths and closest training views, form our training triplets. 
Specifically, DualDiff / DualDiff-s denotes weights trained on results from a single view count (24 views), while DualDiff-m represents weights trained on results from multiple view counts (3, 6, 9, and 24 views). 
We adopt the Adam optimizer with a learning rate of $2\times 10^{-5}$ and a batch size of $8$, training for $10k$ iterations.

\paragraph{DualDiff3D.}
In the 3DGS reconstruction process, we initialize the model with a randomly distributed point cloud to ensure a fair comparison. 
Following the setup in DIFIX3D, the 3DGS is first trained on sparse input views for 30k iterations. 
After the initial 3DGS reconstruction, we run another 30k iterations with a refinement step every 2k iterations. 
For the validation and rollback mechanism, we cache the Gaussian scene state at each refinement step. 
During the 2k iterations of training after caching, we evaluate the model on a held-out validation set every 500 iterations. 
If performance fails to improve for 3 consecutive validations, we roll back to the cached state, discard the newly added refined views, and resume 3DGS training with only the original real image set until the next refinement step. In our experiments, we set the number of refinement steps $K=3$, and loss weights $\lambda_1=0.2, \lambda_2=0.3, \lambda_3=0.5$ via grid parameter search.

\begin{table}[t]
\centering
\setlength\tabcolsep{5pt}
\renewcommand{\arraystretch}{1.15}
\captionsetup{type=table}
\caption{\textbf{Objective refinement results of DualDiff on the DL3DV dataset.}}
\label{tab:refine}
\resizebox{\linewidth}{!}{
\begin{tabular}{llcccclcccc}
\toprule
\textbf{Method} &
\textbf{Setting} & \textbf{PSNR$\uparrow$} & \textbf{SSIM$\uparrow$} & \textbf{FID$\downarrow$} & \textbf{LPIPS$\downarrow$} &
\textbf{Setting} & \textbf{PSNR$\uparrow$} & \textbf{SSIM$\uparrow$} & \textbf{FID$\downarrow$} & \textbf{LPIPS$\downarrow$} \\
\midrule
3DGS &
\multirow{4}{*}{3-View}  & 10.70 & 0.2608 & 267.77 & 0.6001 &
\multirow{4}{*}{6-View}  & 12.72 & 0.3298 & 214.26 & 0.5419 \\
DIFIX &
  & 13.11 & 0.3338 & 133.77 & 0.4966 &
  & 13.99 & 0.3641 & 96.87  & 0.4620 \\
\gc DualDiff-m &
  & \gc \textbf{13.89} & \gc \textbf{0.3793} & \gc 103.67 & \gc \textbf{0.4407} &
  & \gc \textbf{14.67} & \gc \textbf{0.4150} & \gc 66.93 & \gc \textbf{0.4020} \\
\gc DualDiff-s &
  & \gc 13.49 & \gc 0.3674 & \gc \textbf{101.80} & \gc 0.4522 &
  & \gc 14.48 & \gc 0.4086 & \gc \textbf{62.41}  & \gc 0.4060 \\
\midrule
3DGS &
\multirow{4}{*}{9-View}  & 14.19 & 0.3886 & 192.17 & 0.4950 &
\multirow{4}{*}{24-View} & 19.42 & 0.6209 & 98.71  & 0.3212 \\
DIFIX &
 & 16.02 & 0.4399 & 86.55  & 0.3800 &
 & 19.20 & 0.5612 & 46.52  & 0.2730 \\
\gc DualDiff-m &
 & \gc 16.76 & \gc 0.4973 & \gc 60.72 & \gc 0.3088 &
 & \gc 20.27 & \gc 0.6329 & \gc 26.62 & \gc 0.1989 \\
\gc DualDiff-s &
 & \gc \textbf{16.77} & \gc \textbf{0.5001} & \gc \textbf{58.19} & \gc \textbf{0.3079} &
 & \gc \textbf{20.57} & \gc \textbf{0.6456} & \gc \textbf{24.77} & \gc \textbf{0.1931} \\
\bottomrule
\end{tabular}
}
\end{table}

\subsection{Refinement Results}
We evaluate the novel view refinement results of DualDiff, guided by reference views on the remaining 10\% of scenes from the DL3DV dataset, as well as the LLFF dataset, with comparisons against vanilla 3DGS and DIFIX (the refinement model of DIFIX3D+).  
Consistent with the training dataset construction process, we conduct tests on the DL3DV dataset by refining novel views rendered from 3D reconstructions using 3, 6, 9, and 24 sparse input views. 
As shown in \cref{tab:refine}, DualDiff outperforms both vanilla 3DGS and DIFIX by a large margin across all experimental settings.
DualDiff-m demonstrates superior performance in few-view test scenarios, as it has encountered more severe rendering degradations in the training phase. 
Although DualDiff-s is only trained on 24-view reconstructions, its refinement capability remains effective across various test settings. 
The subjective results, as shown in \cref{fig:teaser}, demonstrate that DualDiff significantly reduces artifacts, restores finer details, and achieves the best appearance consistency with the reference view. 
Due to space constraints, additional results are provided in the supplementary material.

\begin{figure*}[t]
    \centering
    \includegraphics[width=1.0\linewidth]{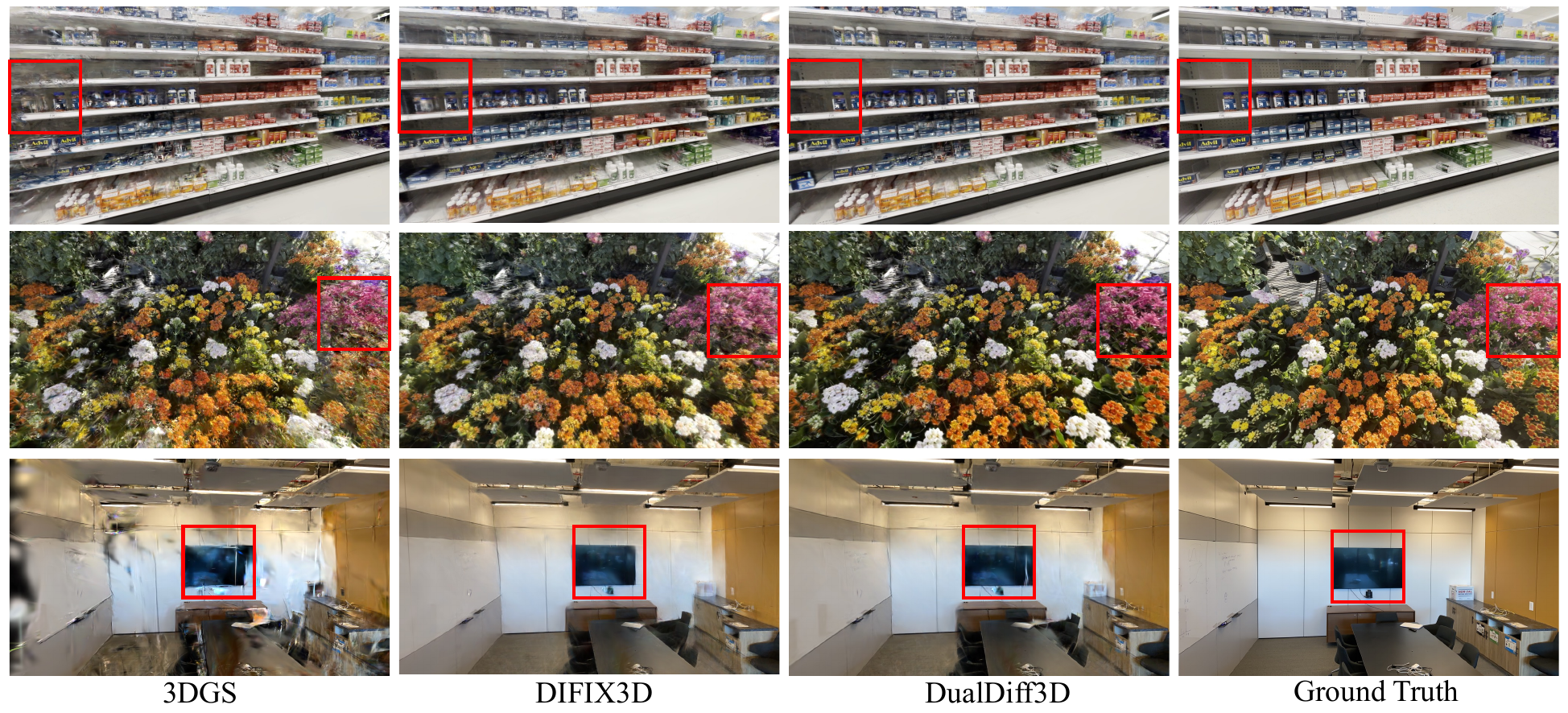}
    \captionsetup{type=figure}
    \caption{\textbf{Comparison of subjective results of DualDiff3D and baselines on the DL3DV and LLFF datasets.} Regions with significant differences are highlighted by red boxes for easy comparison.}
    \label{fig:recon}
\end{figure*}

\begin{table*}[t]
\centering
\setlength\tabcolsep{5pt}
\renewcommand{\arraystretch}{1.15}
\captionsetup{type=table}
\caption{\textbf{Objective reconstruction results of DualDiff3D on the DL3DV and LLFF datasets.}}
\label{tab:recon}
\resizebox{\linewidth}{!}{
\begin{tabular}{lcccccccccccc}
\toprule
\multirow{2}{*}{Method} &
  \multicolumn{3}{c}{3-View} &
  \multicolumn{3}{c}{6-View} &
  \multicolumn{3}{c}{9-View} &
  \multicolumn{3}{c}{24-View} \\
\cmidrule(lr){2-4} \cmidrule(lr){5-7} \cmidrule(lr){8-10} \cmidrule(lr){11-13}
 &
  PSNR$\uparrow$ & SSIM$\uparrow$ & LPIPS$\downarrow$ &
  PSNR$\uparrow$ & SSIM$\uparrow$ & LPIPS$\downarrow$ &
  PSNR$\uparrow$ & SSIM$\uparrow$ & LPIPS$\downarrow$ &
  PSNR$\uparrow$ & SSIM$\uparrow$ & LPIPS$\downarrow$ \\
\midrule
\multicolumn{13}{c}{DL3DV} \\
\midrule
3DGS &
  10.70 & 0.2608 & 0.6001 &
  12.72 & 0.3298 & 0.5419 &
  14.19 & 0.3886 & 0.4950 &
  19.42 & 0.6209 & 0.3212 \\
DIFIX3D &
  11.36 & 0.3314 & 0.5759 &
  13.35 & 0.3892 & 0.5084 &
  14.96 & 0.4464 & 0.4617 &
  20.17 & 0.6469 & 0.2971 \\
\rowcolor[HTML]{F5F5F5}
DualDiff3D &
  \textbf{12.56} & \textbf{0.3731} & \textbf{0.5473} &
  \textbf{14.11} & \textbf{0.4122} & \textbf{0.4837} &
  \textbf{15.60} & \textbf{0.4697} & \textbf{0.4436} &
  \textbf{20.49} & \textbf{0.6576} & \textbf{0.2745} \\
\midrule
\multicolumn{13}{c}{LLFF} \\
\midrule
3DGS &
  14.11 & 0.3819 & 0.4666 &
  18.69 & 0.5687 & 0.3237 &
  20.50 & 0.6508 & 0.2726 &
  - & - & - \\
DIFIX3D &
  15.21 & 0.4645 & 0.4119 &
  19.85 & 0.6159 & 0.2821 &
  21.53 & 0.6807 & 0.2410 &
  - & - & - \\
\rowcolor[HTML]{F5F5F5}
DualDiff3D &
  \textbf{16.34} & \textbf{0.4981} & \textbf{0.3835} &
  \textbf{20.35} & \textbf{0.6709} & \textbf{0.2611} &
  \textbf{21.89} & \textbf{0.7106} & \textbf{0.2154} &
  - & - & - \\
\bottomrule
\end{tabular}
}
\end{table*}

\subsection{Reconstruction Results}
We compare the 3D reconstruction performance of vanilla 3DGS, DIFIX3D, and DualDiff3D on both the DL3DV dataset and the LLFF dataset. \cref{tab:recon} summarizes the objective performance across key metrics, where DualDiff3D achieves the highest metric values under all view count conditions. 
This not only validates the superior reconstruction capability of DualDiff3D but also demonstrates its strong generalization ability considering it was not trained on the LLFF dataset for refinement. 
\cref{fig:recon} visualizes the 3D reconstruction results of the mentioned methods. 
These visual results confirm that DualDiff3D achieves an excellent balance between photorealism and geometric coherence.

\begin{table*}[t]
\centering
\setlength\tabcolsep{5pt}
\renewcommand{\arraystretch}{1.15}
\captionsetup{type=table}
\caption{\textbf{Ablation results for key components of DualDiff3D.}}
\label{tab:ablation}
\resizebox{\linewidth}{!}{
\begin{tabular}{ccccccccccc}
\toprule
\multirow{2}{*}{Variants} & \multicolumn{2}{c}{Components} & \multicolumn{4}{c}{DualDiff} & \multicolumn{4}{c}{DIFIX} \\
\cmidrule(lr){2-3} \cmidrule(lr){4-7} \cmidrule(lr){8-11}
& PSF & CDW & PSNR$\uparrow$ & SSIM$\uparrow$ & LPIPS$\downarrow$ & FID$\downarrow$ & PSNR$\uparrow$ & SSIM$\uparrow$ & LPIPS$\downarrow$ & FID$\downarrow$ \\
\midrule
(a) &            &            & 15.43 & 0.4649 & 0.4031 & 107.50 & 15.21 & 0.4645 & 0.4119 & 115.23 \\
(b) & \checkmark &            & 15.71 & 0.4733 & 0.3910 & 105.02 & 15.44 & 0.4699 & 0.4100 & 111.36 \\
(c) &            & \checkmark & 16.04 & 0.4877 & 0.3881 & 103.71 & 15.79 & 0.4764 & 0.4012 & 108.56 \\
(d) & \checkmark & \checkmark & \textbf{16.34} & \textbf{0.4981} & \textbf{0.3835} & \textbf{101.25} & \textbf{15.96} & \textbf{0.4811} & \textbf{0.3903} & \textbf{106.77} \\
\bottomrule
\end{tabular}}
\end{table*}

Following the evaluation protocol of GenFusion~\cite{genfusion} and GSFixer~\cite{gsfixer}, we compare DualDiff3D with more baselines on Mip-NeRF360~\cite{mipnerf360} (\cref{tab:mipnerf360}). 
Due to the lack of open-source code for 3DGS-Enhancer, we utilize the values reported in their paper. 

\begin{table}[htbp]
\centering
\setlength\tabcolsep{5pt}
\renewcommand{\arraystretch}{1.15}
\captionsetup{type=table}
\caption{\textbf{Comparison with baselines on the Mip-NeRF360 dataset.} $^{*}$ denotes values reported in the original paper.}
\label{tab:mipnerf360}
\resizebox{\linewidth}{!}{
\begin{tabular}{lccccccccc}
\toprule
\multirow{2}{*}{\textbf{Method}} & \multicolumn{3}{c}{\textbf{3-View}} & \multicolumn{3}{c}{\textbf{6-View}} & \multicolumn{3}{c}{\textbf{9-View}} \\
\cmidrule(lr){2-4} \cmidrule(lr){5-7} \cmidrule(lr){8-10}
& PSNR$\uparrow$ & SSIM$\uparrow$ & LPIPS$\downarrow$ & PSNR$\uparrow$ & SSIM$\uparrow$ & LPIPS$\downarrow$ & PSNR$\uparrow$ & SSIM$\uparrow$ & LPIPS$\downarrow$ \\
\midrule
3DGS & 13.06 & 0.251 & 0.576 & 14.96 & 0.355 & 0.505 & 16.79 & 0.447 & 0.446 \\
FSGS & 14.17 & 0.318 & 0.578 & 16.12 & 0.415 & 0.517 & 17.94 & 0.492 & 0.468 \\
GenFusion & 15.29 & 0.369 & 0.585 & 17.16 & 0.447 & 0.500 & 18.36 & 0.496 & 0.465 \\
3DGS-Enhancer$^{*}$ & - & - & - & 13.96 & 0.260 & 0.570 & 16.22 & 0.399 & 0.454 \\
DIFIX3D & 14.32 & 0.313 & 0.571 & 16.07 & 0.408 & 0.459 & 17.82 & 0.477 & 0.418 \\
GSFixer & 15.61 & 0.370 & 0.559 & \textbf{17.27} & 0.426 & 0.478 & \textbf{18.63} & 0.481 & 0.420 \\
\midrule
\rowcolor[HTML]{F5F5F5}
\textbf{Ours} & \textbf{15.70} & \textbf{0.373} & \textbf{0.554} & 17.22 & \textbf{0.439} & \textbf{0.451} & 18.45 & \textbf{0.486} & \textbf{0.405} \\
\bottomrule
\end{tabular}}
\end{table}

\subsection{Ablation Study}
To validate the contributions of key components in DualDiff3D, we conducted ablation studies on the LLFF dataset under the 3-view setting, with quantitative results summarized in \cref{tab:ablation}. 
Specifically, we ablate three key components one by one: the refinement model (replacing DualDiff with DIFIX), the Progressive Sampling and Filtering (PSF) strategy, and the Confidence-Driven Weighting (CDW) process. 
Here, the baseline refers to directly incorporating refined novel views into 3DGS reconstruction without additional optimization. 
The full DualDiff3D model integrates all components and delivers the best performance. We also ablate $K$ and each CDW term in \cref{tab:cdw-ablation}. $K\!=\!3$ achieves a quality/cost trade-off, and combining three terms yields the best performance.

\begin{table}[htbp]
\centering
\setlength\tabcolsep{5pt}
\renewcommand{\arraystretch}{1.15}
\captionsetup{type=table}
\caption{\textbf{Ablation on $K$ and individual CDW terms.}}
\label{tab:cdw-ablation}
\resizebox{\linewidth}{!}{
\begin{tabular}{lccccccc}
\toprule
\textbf{Variant} & $K$ & $C_{\text{var}}$ & $C_{\text{diff}}$ & $C_{\text{proj}}$ & \textbf{PSNR$\uparrow$} & \textbf{SSIM$\uparrow$} & \textbf{LPIPS$\downarrow$} \\
\midrule
w/o CDW               & 1 &            &            &            & 19.97 & 0.6213 & 0.2704 \\
only $C_{\text{var}}$  & 3 & \checkmark &            &            & 20.06 & 0.6225 & 0.2715 \\
only $C_{\text{diff}}$ & 3 &            & \checkmark &            & 20.11 & 0.6310 & 0.2696 \\
only $C_{\text{proj}}$ & 3 &            &            & \checkmark & 20.26 & 0.6612 & 0.2633 \\
\rowcolor[HTML]{F5F5F5}
\textbf{w/ full CDW (Ours)} & 3 & \checkmark & \checkmark & \checkmark & \textbf{20.35} & \textbf{0.6709} & \textbf{0.2611} \\
\midrule
w/ full CDW           & 2 & \checkmark & \checkmark & \checkmark & 20.26 & 0.6633 & 0.2626 \\
w/ full CDW           & 4 & \checkmark & \checkmark & \checkmark & 20.37 & 0.6701 & 0.2609 \\
w/ full CDW           & 5 & \checkmark & \checkmark & \checkmark & 20.33 & 0.6689 & 0.2606 \\
\bottomrule
\end{tabular}}
\end{table}

\subsection{Computational Cost}
We report the runtime parameters of three methods in \cref{tab:runtime}. 
Despite having a larger backbone network, DualDiff3D outperforms DIFIX3D in terms of average single refining runtime. 
This is primarily because DIFIX3D’s approach of placing reference views in the batch dimension restricts the refinement of only one input view at a time, whereas our method employs two separate networks to process input views and reference views, respectively. 
With refined times $k=3$ for reconstruction, our final runtime is slightly longer. 
However, considering the substantial performance gains and the fact that our method is not designed for real-time 3DGS reconstruction, we argue that this time overhead is justifiable.

\begin{table}[htbp]
\centering
\setlength\tabcolsep{5pt}
\renewcommand{\arraystretch}{1.15}
\captionsetup{type=table}
\caption{\textbf{Runtime parameters on a single RTX 4090 GPU.}}
\label{tab:runtime}
\resizebox{\linewidth}{!}{
\begin{tabular}{lcccc}
\toprule
 & RefineTime & Recon.~Time & RefineVRAM-Avg & RefineVRAM-Peak \\
\midrule
3DGS       & -     & 6min29s  & -        & -         \\
DIFIX3D    & 950ms & 8min52s  & 6048.2MB & 9936.7MB  \\
\rowcolor[HTML]{F5F5F5}
DualDiff3D & 600ms & 10min40s & 9750.5MB & 12574.6MB \\
\bottomrule
\end{tabular}}
\end{table}
\section{Conclusion}
In this paper, we present DualDiff3D, a framework designed to advance 3D reconstruction and novel-view synthesis. 
At the core of our framework lies DualDiff, a dual-branch diffusion pipeline tailored to eliminate artifacts in novel views rendered from 3DGS. 
A proposed SAA mechanism connects the structure branch and appearance branch, enabling image quality improvement in a ``free lunch'' manner. 
DualDiff3D leverages a Render-Refine-Optimize Loop, integrated with sampling strategy and confidence-guided masks, to progressively and robustly refine 3DGS. 
Extensive experiments validate the effectiveness of our approach, which delivers superior visual quality compared to baselines.
We hope that DualDiff3D offers a more rational solution for diffusion-based NVS and 3D reconstruction, and may inspire more effective designs.

\subsection{Limitation and Future Work}
Despite being a one-step diffusion architecture with fast inference speed, DualDiff3D still imposes overhead in computational resources. 
Furthermore, in extremely sparse-view scenarios, the model’s performance may suffer significant degradation.
Thus, future work may focus on model distillation and extending it to extremely sparse input scenarios.
\section{Acknowledgements}
This work is financially supported in part by National Natural Science Foundation of China (62372016), National Science and Technology Major Project-Mobile Information Networks (2024ZD130060), Guangdong Provincial Key Laboratory of Ultra High Definition Immersive Media Technology (2024B1212010006), Shenzhen Science and Technology Program (SYSPG20241211173440004) and Outstanding Talents Training Fund in Shenzhen.
\bibliographystyle{splncs04}
\bibliography{main}

@String(CVPR= {IEEE Conf. Comput. Vis. Pattern Recog.})

@String(TOG= {ACM Trans. Graph.})

@String(ICLR = {Int. Conf. Learn. Represent.})

@String(AAAI = {AAAI})

@String(CVPR  = {CVPR})

@String(TOG   = {ACM TOG})

@String(ICLR  = {ICLR})

@article{nerf,
  title={Nerf: Representing scenes as neural radiance fields for view synthesis},
  author={Mildenhall, Ben and Srinivasan, Pratul P and Tancik, Matthew and Barron, Jonathan T and Ramamoorthi, Ravi and Ng, Ren},
  journal={Communications of the ACM},
  volume={65},
  number={1},
  pages={99--106},
  year={2021},
  publisher={ACM New York, NY, USA}
}

@article{3dgs,
  title={3D Gaussian splatting for real-time radiance field rendering.},
  author={Kerbl, Bernhard and Kopanas, Georgios and Leimk{\"u}hler, Thomas and Drettakis, George},
  journal={ACM Trans. Graph.},
  volume={42},
  number={4},
  pages={139--1},
  year={2023}
}

@inproceedings{difix3d,
  title={Difix3d+: Improving 3d reconstructions with single-step diffusion models},
  author={Wu, Jay Zhangjie and Zhang, Yuxuan and Turki, Haithem and Ren, Xuanchi and Gao, Jun and Shou, Mike Zheng and Fidler, Sanja and Gojcic, Zan and Ling, Huan},
  booktitle={Proceedings of the Computer Vision and Pattern Recognition Conference},
  pages={26024--26035},
  year={2025}
}

@article{3dgsenhancer,
  title={3dgs-enhancer: Enhancing unbounded 3d gaussian splatting with view-consistent 2d diffusion priors},
  author={Liu, Xi and Zhou, Chaoyi and Huang, Siyu},
  journal={Advances in Neural Information Processing Systems},
  volume={37},
  pages={133305--133327},
  year={2024}
}

@article{nerfactor,
  title={Nerfactor: Neural factorization of shape and reflectance under an unknown illumination},
  author={Zhang, Xiuming and Srinivasan, Pratul P and Deng, Boyang and Debevec, Paul and Freeman, William T and Barron, Jonathan T},
  journal={ACM Transactions on Graphics (ToG)},
  volume={40},
  number={6},
  pages={1--18},
  year={2021},
  publisher={ACM New York, NY, USA}
}

@inproceedings{zipnerf,
  title={Zip-nerf: Anti-aliased grid-based neural radiance fields},
  author={Barron, Jonathan T and Mildenhall, Ben and Verbin, Dor and Srinivasan, Pratul P and Hedman, Peter},
  booktitle={Proceedings of the IEEE/CVF International Conference on Computer Vision},
  pages={19697--19705},
  year={2023}
}

@inproceedings{fsgs,
  title={Fsgs: Real-time few-shot view synthesis using gaussian splatting},
  author={Zhu, Zehao and Fan, Zhiwen and Jiang, Yifan and Wang, Zhangyang},
  booktitle={European conference on computer vision},
  pages={145--163},
  year={2024},
  organization={Springer}
}

@article{ddim,
  title={Denoising diffusion implicit models},
  author={Song, Jiaming and Meng, Chenlin and Ermon, Stefano},
  journal={arXiv preprint arXiv:2010.02502},
  year={2020}
}

@article{ddpm,
  title={Denoising diffusion probabilistic models},
  author={Ho, Jonathan and Jain, Ajay and Abbeel, Pieter},
  journal={Advances in neural information processing systems},
  volume={33},
  pages={6840--6851},
  year={2020}
}

@article{gan,
  title={Generative adversarial nets},
  author={Goodfellow, Ian J and Pouget-Abadie, Jean and Mirza, Mehdi and Xu, Bing and Warde-Farley, David and Ozair, Sherjil and Courville, Aaron and Bengio, Yoshua},
  journal={Advances in neural information processing systems},
  volume={27},
  year={2014}
}

@inproceedings{stackgan,
  title={Stackgan: Text to photo-realistic image synthesis with stacked generative adversarial networks},
  author={Zhang, Han and Xu, Tao and Li, Hongsheng and Zhang, Shaoting and Wang, Xiaogang and Huang, Xiaolei and Metaxas, Dimitris N},
  booktitle={Proceedings of the IEEE international conference on computer vision},
  pages={5907--5915},
  year={2017}
}

@article{vae,
  title={Auto-encoding variational bayes},
  author={Kingma, Diederik P and Welling, Max},
  journal={arXiv preprint arXiv:1312.6114},
  year={2013}
}

@article{vqvae,
  title={Neural discrete representation learning},
  author={Van Den Oord, Aaron and Vinyals, Oriol and others},
  journal={Advances in neural information processing systems},
  volume={30},
  year={2017}
}

@article{flow,
  title={Flow matching for generative modeling},
  author={Lipman, Yaron and Chen, Ricky TQ and Ben-Hamu, Heli and Nickel, Maximilian and Le, Matt},
  journal={arXiv preprint arXiv:2210.02747},
  year={2022}
}

@article{glide,
  title={Glide: Towards photorealistic image generation and editing with text-guided diffusion models},
  author={Nichol, Alex and Dhariwal, Prafulla and Ramesh, Aditya and Shyam, Pranav and Mishkin, Pamela and McGrew, Bob and Sutskever, Ilya and Chen, Mark},
  journal={arXiv preprint arXiv:2112.10741},
  year={2021}
}

@article{cfg,
  title={Classifier-free diffusion guidance},
  author={Ho, Jonathan and Salimans, Tim},
  journal={arXiv preprint arXiv:2207.12598},
  year={2022}
}

@inproceedings{clip,
  title={Learning transferable visual models from natural language supervision},
  author={Radford, Alec and Kim, Jong Wook and Hallacy, Chris and Ramesh, Aditya and Goh, Gabriel and Agarwal, Sandhini and Sastry, Girish and Askell, Amanda and Mishkin, Pamela and Clark, Jack and others},
  booktitle={International conference on machine learning},
  pages={8748--8763},
  year={2021},
  organization={PmLR}
}

@article{dalle2,
  title={Hierarchical text-conditional image generation with clip latents},
  author={Ramesh, Aditya and Dhariwal, Prafulla and Nichol, Alex and Chu, Casey and Chen, Mark},
  journal={arXiv preprint arXiv:2204.06125},
  volume={1},
  number={2},
  pages={3},
  year={2022}
}

@inproceedings{KinectFusion,
  title={Kinectfusion: Real-time dense surface mapping and tracking},
  author={Newcombe, Richard A and Izadi, Shahram and Hilliges, Otmar and Molyneaux, David and Kim, David and Davison, Andrew J and Kohi, Pushmeet and Shotton, Jamie and Hodges, Steve and Fitzgibbon, Andrew},
  booktitle={2011 10th IEEE international symposium on mixed and augmented reality},
  pages={127--136},
  year={2011},
  organization={Ieee}
}

@article{imagen,
  title={Photorealistic text-to-image diffusion models with deep language understanding},
  author={Saharia, Chitwan and Chan, William and Saxena, Saurabh and Li, Lala and Whang, Jay and Denton, Emily L and Ghasemipour, Kamyar and Gontijo Lopes, Raphael and Karagol Ayan, Burcu and Salimans, Tim and others},
  journal={Advances in neural information processing systems},
  volume={35},
  pages={36479--36494},
  year={2022}
}

@inproceedings{ldm,
  title={High-resolution image synthesis with latent diffusion models},
  author={Rombach, Robin and Blattmann, Andreas and Lorenz, Dominik and Esser, Patrick and Ommer, Bj{\"o}rn},
  booktitle={Proceedings of the IEEE/CVF conference on computer vision and pattern recognition},
  pages={10684--10695},
  year={2022}
}

@inproceedings{dreambooth,
  title={Dreambooth: Fine tuning text-to-image diffusion models for subject-driven generation},
  author={Ruiz, Nataniel and Li, Yuanzhen and Jampani, Varun and Pritch, Yael and Rubinstein, Michael and Aberman, Kfir},
  booktitle={Proceedings of the IEEE/CVF conference on computer vision and pattern recognition},
  pages={22500--22510},
  year={2023}
}

@article{textinversion,
  title={An image is worth one word: Personalizing text-to-image generation using textual inversion},
  author={Gal, Rinon and Alaluf, Yuval and Atzmon, Yuval and Patashnik, Or and Bermano, Amit H and Chechik, Gal and Cohen-Or, Daniel},
  journal={arXiv preprint arXiv:2208.01618},
  year={2022}
}

@inproceedings{paint,
  title={Paint by example: Exemplar-based image editing with diffusion models},
  author={Yang, Binxin and Gu, Shuyang and Zhang, Bo and Zhang, Ting and Chen, Xuejin and Sun, Xiaoyan and Chen, Dong and Wen, Fang},
  booktitle={Proceedings of the IEEE/CVF conference on computer vision and pattern recognition},
  pages={18381--18391},
  year={2023}
}

@inproceedings{t2iadapter,
  title={T2i-adapter: Learning adapters to dig out more controllable ability for text-to-image diffusion models},
  author={Mou, Chong and Wang, Xintao and Xie, Liangbin and Wu, Yanze and Zhang, Jian and Qi, Zhongang and Shan, Ying},
  booktitle={Proceedings of the AAAI conference on artificial intelligence},
  volume={38},
  number={5},
  pages={4296--4304},
  year={2024}
}

@inproceedings{controlnet,
  title={Adding conditional control to text-to-image diffusion models},
  author={Zhang, Lvmin and Rao, Anyi and Agrawala, Maneesh},
  booktitle={Proceedings of the IEEE/CVF international conference on computer vision},
  pages={3836--3847},
  year={2023}
}

@inproceedings{mipnerf,
  title={Mip-nerf: A multiscale representation for anti-aliasing neural radiance fields},
  author={Barron, Jonathan T and Mildenhall, Ben and Tancik, Matthew and Hedman, Peter and Martin-Brualla, Ricardo and Srinivasan, Pratul P},
  booktitle={Proceedings of the IEEE/CVF international conference on computer vision},
  pages={5855--5864},
  year={2021}
}

@inproceedings{tensorf,
  title={Tensorf: Tensorial radiance fields},
  author={Chen, Anpei and Xu, Zexiang and Geiger, Andreas and Yu, Jingyi and Su, Hao},
  booktitle={European conference on computer vision},
  pages={333--350},
  year={2022},
  organization={Springer}
}

@inproceedings{plenoxels,
  title={Plenoxels: Radiance fields without neural networks},
  author={Fridovich-Keil, Sara and Yu, Alex and Tancik, Matthew and Chen, Qinhong and Recht, Benjamin and Kanazawa, Angjoo},
  booktitle={Proceedings of the IEEE/CVF conference on computer vision and pattern recognition},
  pages={5501--5510},
  year={2022}
}

@article{instantngp,
  title={Instant neural graphics primitives with a multiresolution hash encoding},
  author={M{\"u}ller, Thomas and Evans, Alex and Schied, Christoph and Keller, Alexander},
  journal={ACM transactions on graphics (TOG)},
  volume={41},
  number={4},
  pages={1--15},
  year={2022},
  publisher={ACM New York, NY, USA}
}

@inproceedings{occupancy,
  title={Occupancy networks: Learning 3d reconstruction in function space},
  author={Mescheder, Lars and Oechsle, Michael and Niemeyer, Michael and Nowozin, Sebastian and Geiger, Andreas},
  booktitle={Proceedings of the IEEE/CVF conference on computer vision and pattern recognition},
  pages={4460--4470},
  year={2019}
}

@inproceedings{deepsdf,
  title={Deepsdf: Learning continuous signed distance functions for shape representation},
  author={Park, Jeong Joon and Florence, Peter and Straub, Julian and Newcombe, Richard and Lovegrove, Steven},
  booktitle={Proceedings of the IEEE/CVF conference on computer vision and pattern recognition},
  pages={165--174},
  year={2019}
}

@article{gaussiansr,
  title={Gaussiansr: 3d gaussian super-resolution with 2d diffusion priors},
  author={Yu, Xiqian and Zhu, Hanxin and He, Tianyu and Chen, Zhibo},
  journal={arXiv preprint arXiv:2406.10111},
  year={2024}
}

@article{diffusionprior,
  title={Diffusion priors for dynamic view synthesis from monocular videos},
  author={Wang, Chaoyang and Zhuang, Peiye and Siarohin, Aliaksandr and Cao, Junli and Qian, Guocheng and Lee, Hsin-Ying and Tulyakov, Sergey},
  journal={arXiv preprint arXiv:2401.05583},
  year={2024}
}

@article{dreamsparse,
  title={Dreamsparse: Escaping from plato’s cave with 2d diffusion model given sparse views},
  author={Yoo, Paul and Guo, Jiaxian and Matsuo, Yutaka and Gu, Shixiang Shane},
  journal={Advances in Neural Information Processing Systems},
  volume={36},
  pages={3307--3324},
  year={2023}
}

@article{yifan2019differentiable,
  title={Differentiable surface splatting for point-based geometry processing},
  author={Yifan, Wang and Serena, Felice and Wu, Shihao and {\"O}ztireli, Cengiz and Sorkine-Hornung, Olga},
  journal={ACM Transactions On Graphics (TOG)},
  volume={38},
  number={6},
  pages={1--14},
  year={2019},
  publisher={ACM New York, NY, USA}
}

@article{kopanas2022neural,
  title={Neural point catacaustics for novel-view synthesis of reflections},
  author={Kopanas, Georgios and Leimk{\"u}hler, Thomas and Rainer, Gilles and Jambon, Cl{\'e}ment and Drettakis, George},
  journal={ACM Transactions on Graphics (TOG)},
  volume={41},
  number={6},
  pages={1--15},
  year={2022},
  publisher={ACM New York, NY, USA}
}

@article{zwicker2002ewa,
  title={EWA splatting},
  author={Zwicker, Matthias and Pfister, Hanspeter and Van Baar, Jeroen and Gross, Markus},
  journal={IEEE Transactions on Visualization and Computer Graphics},
  volume={8},
  number={3},
  pages={223--238},
  year={2002},
  publisher={IEEE}
}

@article{ipadapter,
  title={Ip-adapter: Text compatible image prompt adapter for text-to-image diffusion models},
  author={Ye, Hu and Zhang, Jun and Liu, Sibo and Han, Xiao and Yang, Wei},
  journal={arXiv preprint arXiv:2308.06721},
  year={2023}
}

@inproceedings{animateanyone,
  title={Animate anyone: Consistent and controllable image-to-video synthesis for character animation},
  author={Hu, Li},
  booktitle={Proceedings of the IEEE/CVF Conference on Computer Vision and Pattern Recognition},
  pages={8153--8163},
  year={2024}
}

@inproceedings{sdturbo,
  title={Fast high-resolution image synthesis with latent adversarial diffusion distillation},
  author={Sauer, Axel and Boesel, Frederic and Dockhorn, Tim and Blattmann, Andreas and Esser, Patrick and Rombach, Robin},
  booktitle={SIGGRAPH Asia 2024 Conference Papers},
  pages={1--11},
  year={2024}
}

@article{lora,
  title={Lora: Low-rank adaptation of large language models.},
  author={Hu, Edward J and Shen, Yelong and Wallis, Phillip and Allen-Zhu, Zeyuan and Li, Yuanzhi and Wang, Shean and Wang, Lu and Chen, Weizhu and others},
  journal={ICLR},
  volume={1},
  number={2},
  pages={3},
  year={2022}
}

@article{vgg,
  title={Very deep convolutional networks for large-scale image recognition},
  author={Simonyan, Karen and Zisserman, Andrew},
  journal={arXiv preprint arXiv:1409.1556},
  year={2014}
}

@inproceedings{dl3dv,
  title={Dl3dv-10k: A large-scale scene dataset for deep learning-based 3d vision},
  author={Ling, Lu and Sheng, Yichen and Tu, Zhi and Zhao, Wentian and Xin, Cheng and Wan, Kun and Yu, Lantao and Guo, Qianyu and Yu, Zixun and Lu, Yawen and others},
  booktitle={Proceedings of the IEEE/CVF Conference on Computer Vision and Pattern Recognition},
  pages={22160--22169},
  year={2024}
}

@inproceedings{dropgaussian,
  title={Dropgaussian: Structural regularization for sparse-view gaussian splatting},
  author={Park, Hyunwoo and Ryu, Gun and Kim, Wonjun},
  booktitle={Proceedings of the computer vision and pattern recognition conference},
  pages={21600--21609},
  year={2025}
}

@article{llff,
  title={Local light field fusion: Practical view synthesis with prescriptive sampling guidelines},
  author={Mildenhall, Ben and Srinivasan, Pratul P and Ortiz-Cayon, Rodrigo and Kalantari, Nima Khademi and Ramamoorthi, Ravi and Ng, Ren and Kar, Abhishek},
  journal={ACM Transactions on Graphics (ToG)},
  volume={38},
  number={4},
  pages={1--14},
  year={2019},
  publisher={ACM New York, NY, USA}
}

@article{gsfixer,
  title={Gsfixer: Improving 3d gaussian splatting with reference-guided video diffusion priors},
  author={Yin, Xingyilang and Zhang, Qi and Chang, Jiahao and Feng, Ying and Fan, Qingnan and Yang, Xi and Pun, Chi-Man and Zhang, Huaqi and Cun, Xiaodong},
  journal={arXiv preprint arXiv:2508.09667},
  year={2025}
}

@inproceedings{genfusion,
  title={Genfusion: Closing the loop between reconstruction and generation via videos},
  author={Wu, Sibo and Xu, Congrong and Huang, Binbin and Geiger, Andreas and Chen, Anpei},
  booktitle={Proceedings of the Computer Vision and Pattern Recognition Conference},
  pages={6078--6088},
  year={2025}
}

@inproceedings{mipnerf360,
  title={Mip-nerf 360: Unbounded anti-aliased neural radiance fields},
  author={Barron, Jonathan T and Mildenhall, Ben and Verbin, Dor and Srinivasan, Pratul P and Hedman, Peter},
  booktitle={Proceedings of the IEEE/CVF conference on computer vision and pattern recognition},
  pages={5470--5479},
  year={2022}
}

@inproceedings{wang2024360dvd,
  title={360dvd: Controllable panorama video generation with 360-degree video diffusion model},
  author={Wang, Qian and Li, Weiqi and Mou, Chong and Cheng, Xinhua and Zhang, Jian},
  booktitle={2024 IEEE/CVF Conference on Computer Vision and Pattern Recognition (CVPR)},
  pages={6913--6923},
  year={2024},
  organization={IEEE}
}

@article{li2026omnidrag,
  title={Omnidrag: Enabling motion control for omnidirectional image-to-video generation},
  author={Li, Weiqi and Zhao, Shijie and Mou, Chong and Sheng, Xuhan and Zhang, Zhenyu and Wang, Qian and Li, Junlin and Zhang, Li and Zhang, Jian},
  journal={International Journal of Computer Vision},
  volume={134},
  number={1},
  pages={44},
  year={2026},
  publisher={Springer}
}

@article{yang20264dvd,
  title={4DVD: cascaded dense-view video diffusion model for high-quality 4D content generation},
  author={Yang, Shuzhou and Cun, Xiaodong and Li, Xiaoyu and Li, Yaowei and Zhang, Jian},
  journal={International Journal of Computer Vision},
  volume={134},
  number={5},
  pages={233},
  year={2026},
  publisher={Springer}
}

@inproceedings{yang2026gencompositor,
  title={Gencompositor: generative video compositing with diffusion transformer},
  author={Yang, Shuzhou and Li, Xiaoyu and Cun, Xiaodong and Wang, Guangzhi and Li, Lingen and Shan, Ying and Zhang, Jian},
  booktitle={International Conference on Learning Representations},
  volume={2026},
  pages={65477--65501},
  year={2026}
}

@article{li2026q,
  title={Q-insight: Understanding image quality via visual reinforcement learning},
  author={Li, Weiqi and Zhang, Xuanyu and Zhao, Shijie and Zhang, Yabin and Li, Junlin and Zhang, Jian and others},
  journal={Advances in Neural Information Processing Systems},
  volume={38},
  pages={36802--36827},
  year={2026}
}

@inproceedings{li2026uare,
  title={Uare: A unified vision-language model for image quality assessment, restoration, and enhancement},
  author={Li, Weiqi and Zhang, Xuanyu and Chen, Bin and Xie, Jingfen and Wang, Yan and Zhang, Kexin and Li, Junlin and Zhang, Jian and Zhao, Shijie and others},
  booktitle={Proceedings of the IEEE/CVF Conference on Computer Vision and Pattern Recognition},
  pages={22689--22702},
  year={2026}
}

@article{yang2025hybrid,
  title={Hybrid Fourier score distillation for efficient one image to 3D object generation},
  author={Yang, Shuzhou and Wang, Yu and Li, Haijie and Meng, Jiarui and Wu, Yanmin and Meng, Xiandong and Zhang, Jian},
  journal={Visual Intelligence},
  volume={3},
  number={1},
  pages={17},
  year={2025},
  publisher={Springer}
}

@article{cheng2023progressive3d,
  title={Progressive3d: Progressively local editing for text-to-3d content creation with complex semantic prompts},
  author={Cheng, Xinhua and Yang, Tianyu and Wang, Jianan and Li, Yu and Zhang, Lei and Zhang, Jian and Yuan, Li},
  journal={arXiv preprint arXiv:2310.11784},
  year={2023}
}

@inproceedings{zhang2024repaint123,
  title={Repaint123: Fast and high-quality one image to 3d generation with progressive controllable repainting},
  author={Zhang, Junwu and Tang, Zhenyu and Pang, Yatian and Cheng, Xinhua and Jin, Peng and Wei, Yida and Zhou, Xing and Ning, Munan and Yuan, Li},
  booktitle={European Conference on Computer Vision},
  pages={303--320},
  year={2024},
  organization={Springer}
}

@inproceedings{cheng2023panoptic,
  title={Panoptic compositional feature field for editable scene rendering with network-inferred labels via metric learning},
  author={Cheng, Xinhua and Wu, Yanmin and Jia, Mengxi and Wang, Qian and Zhang, Jian},
  booktitle={2023 IEEE/CVF Conference on Computer Vision and Pattern Recognition (CVPR)},
  pages={4947--4957},
  year={2023},
  organization={IEEE}
}

@inproceedings{cheng2026360explorer,
  title={360Explorer: Exploring 4D Controllable World in Panoramic Videos},
  author={Cheng, Xinhua and Zhou, Haiyang and Yu, Wangbo and Jia, Tanghui and Lin, Bin and Ge, Yunyang and Li, Weiqi and Yuan, Li},
  booktitle={Proceedings of the AAAI Conference on Artificial Intelligence},
  volume={40},
  number={5},
  pages={3300--3308},
  year={2026}
}

@inproceedings{zhang2025securegs,
  title={Securegs: Boosting the security and fidelity of 3d gaussian splatting steganography},
  author={Zhang, Xuanyu and Meng, Jiarui and Xu, Zhipei and Yang, Shuzhou and Wu, Yanmin and Wang, Ronggang and Zhang, Jian},
  booktitle={International Conference on Learning Representations},
  volume={2025},
  pages={31654--31673},
  year={2025}
}

@article{zhang2024gs,
  title={Gs-hider: Hiding messages into 3d gaussian splatting},
  author={Zhang, Xuanyu and Meng, Jiarui and Li, Runyi and Xu, Zhipei and Zhang, Yongbing and Zhang, Jian},
  journal={Advances in Neural Information Processing Systems},
  volume={37},
  pages={49780--49805},
  year={2024}
}

@inproceedings{met3r,
  title={Met3r: Measuring multi-view consistency in generated images},
  author={Asim, Mohammad and Wewer, Christopher and Wimmer, Thomas and Schiele, Bernt and Lenssen, Jan Eric},
  booktitle={2025 IEEE/CVF Conference on Computer Vision and Pattern Recognition (CVPR)},
  pages={6034--6044},
  year={2025},
  organization={IEEE}
}

\clearpage
\clearpage
\appendix
\setcounter{page}{1}
\setcounter{figure}{0}
\setcounter{table}{0}
\setcounter{equation}{0}
\renewcommand{\thefigure}{\thesection.\arabic{figure}}
\renewcommand{\thetable}{\thesection.\arabic{table}}
\renewcommand{\theequation}{\thesection.\arabic{equation}}

\section{More Subjective Results}
\begin{figure}[htbp]
    \centering
    \includegraphics[width=\linewidth]{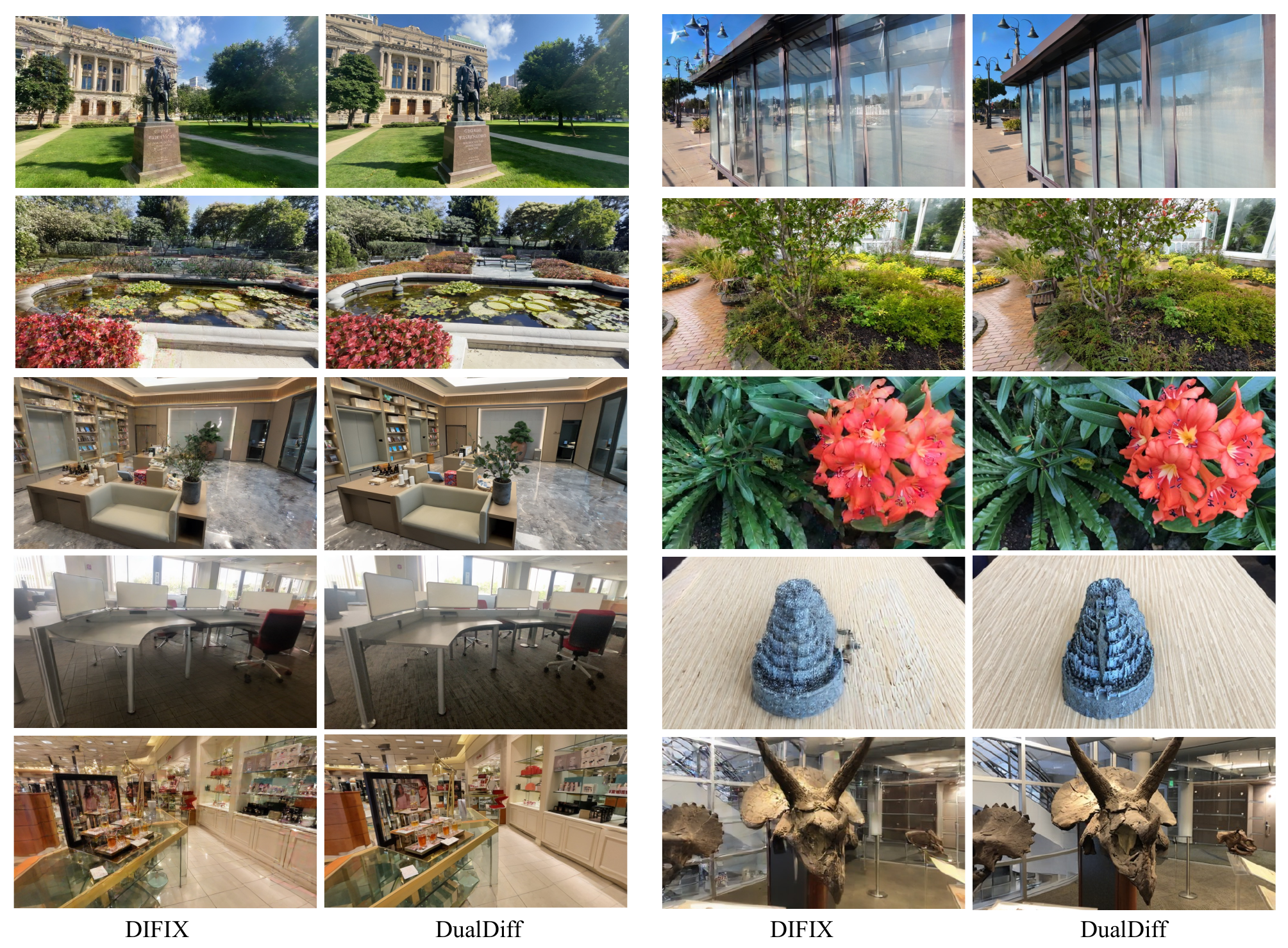}
    \captionsetup{type=figure}
    \caption{\textbf{Comparison of the refinement results of DualDiff and DIFIX~\cite{difix3d} on the DL3DV~\cite{dl3dv} and LLFF~\cite{llff} datasets.}}
    \label{fig:refine-more}
\end{figure}

\begin{figure*}[t]
    \centering
    \includegraphics[width=\linewidth]{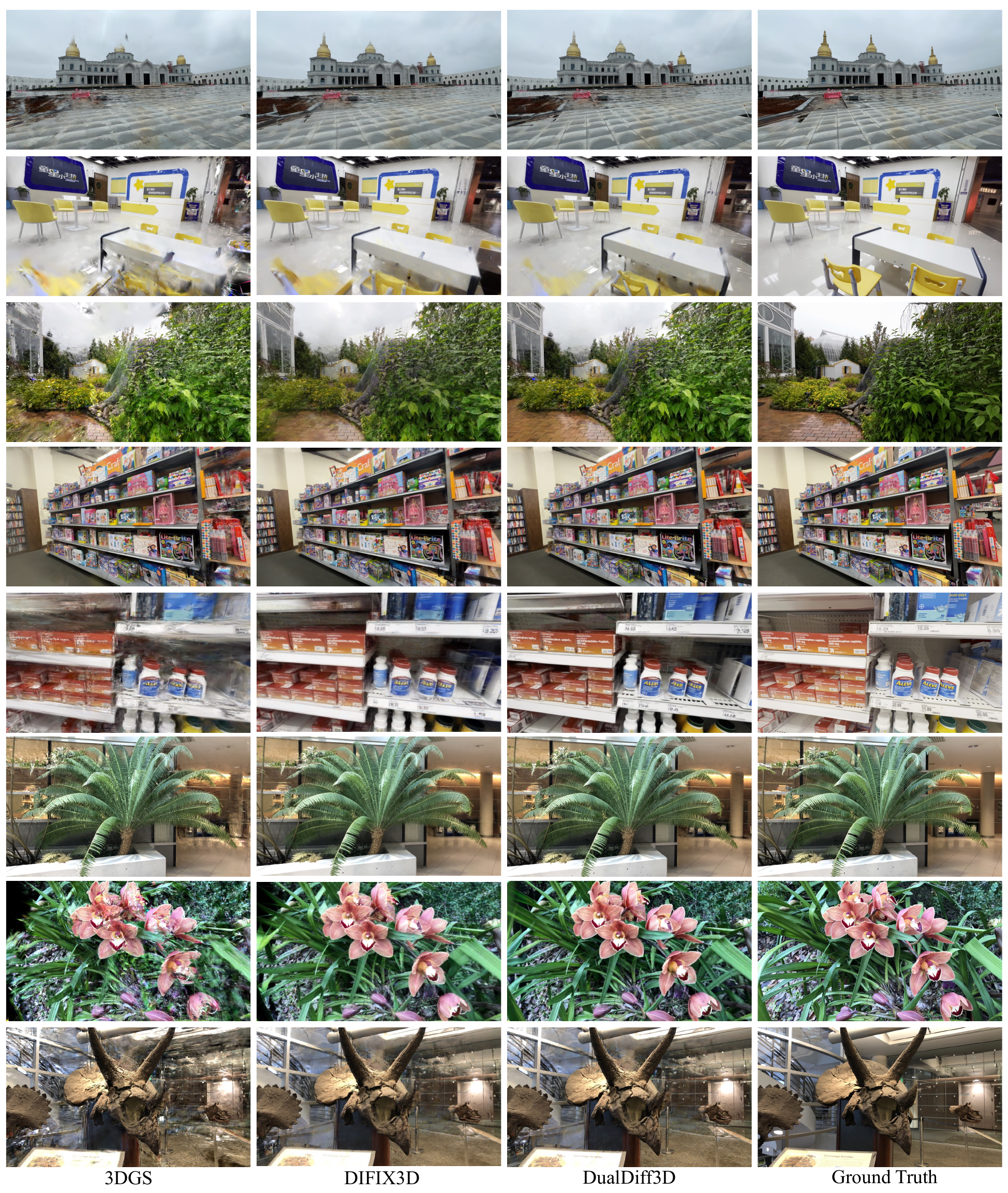}
    \captionsetup{type=figure}
    \caption{\textbf{Comparison of subjective results of DualDiff3D and baselines on DL3DV and LLFF datasets.}}
    \label{fig:recon-more}
\end{figure*}

As shown in \cref{fig:refine-more}, DualDiff outperforms DIFIX~\cite{difix3d} in subjective image quality across various indoor and outdoor scenes in both the DL3DV~\cite{dl3dv} and LLFF~\cite{llff} datasets.
\cref{fig:recon-more} presents additional 3D reconstruction results of DualDiff3D on the DL3DV and LLFF datasets, with comparisons to 3DGS~\cite{3dgs}, DIFIX, and the ground truth. It demonstrates that DualDiff3D effectively enhances reconstruction quality both on the training dataset and out-of-distribution datasets, particularly in edge sharpness and artifact removal.

\begin{table}[htbp]
\centering
\setlength\tabcolsep{5pt}
\renewcommand{\arraystretch}{1.15}
\captionsetup{type=table}
\caption{\textbf{Quantitative comparison of different combinations.} We evaluate DIFIX3D and DualDiff3D with random initialization, 3DGS~\cite{3dgs}, FSGS~\cite{fsgs}, and DropGaussian~\cite{dropgaussian} under sparse-view settings with 3, 6, and 9 input views, respectively.}
\label{tab:generalization}
\resizebox{\linewidth}{!}{
\begin{tabular}{lccccccccc}
\toprule
\multirow{2}{*}{\textbf{Method}} & \multicolumn{3}{c}{\textbf{3-View}} & \multicolumn{3}{c}{\textbf{6-View}} & \multicolumn{3}{c}{\textbf{9-View}} \\
\cmidrule(lr){2-4} \cmidrule(lr){5-7} \cmidrule(lr){8-10}
& \textbf{PSNR$\uparrow$} & \textbf{SSIM$\uparrow$} & \textbf{LPIPS$\downarrow$} & \textbf{PSNR$\uparrow$} & \textbf{SSIM$\uparrow$} & \textbf{LPIPS$\downarrow$} & \textbf{PSNR$\uparrow$} & \textbf{SSIM$\uparrow$} & \textbf{LPIPS$\downarrow$} \\
\midrule
3DGS & 14.11 & 0.3819 & 0.4666 & 18.69 & 0.5687 & 0.3237 & 20.50 & 0.6508 & 0.2726 \\
3DGS+DIFIX3D & 15.21 & 0.4645 & 0.4119 & 19.85 & 0.6159 & 0.2821 & 21.53 & 0.6807 & 0.2410 \\
3DGS+DualDiff3D & 16.34 & 0.4981 & 0.3835 & 20.35 & 0.6709 & 0.2611 & 21.89 & 0.7106 & 0.2154 \\
\midrule
Random+DIFIX3D & 17.85 & 0.5414 & 0.3217 & 21.09 & 0.6548 & 0.2410 & 22.26 & 0.7041 & 0.2128 \\
Random+DualDiff3D & 18.44 & 0.5610 & 0.3135 & 22.64 & 0.7220 & 0.2049 & 23.60 & 0.7971 & 0.1802 \\
\midrule
FSGS & 19.78 & 0.6660 & 0.2667 & 23.21 & 0.7763 & 0.1908 & 24.58 & 0.8159 & 0.1581 \\
FSGS+DIFIX3D & 19.53 & 0.6365 & 0.2650 & 22.75 & 0.7522 & 0.1933 & 24.29 & 0.7984 & 0.1514 \\
\gc FSGS+DualDiff3D & \gc 21.34 & \gc 0.6910 & \gc 0.2504 & \gc 23.93 & \gc 0.7949 & \gc 0.1742 & \gc 24.90 & \gc 0.8373 & \gc 0.1437 \\
\midrule
DropGaussian & 19.65 & 0.6733 & 0.2636 & 23.59 & 0.7873 & 0.1818 & 24.96 & 0.8242 & 0.1492 \\
DropGaussian+DIFIX3D & 22.28 & 0.7469 & 0.1959 & 23.81 & 0.7844 & 0.1796 & 24.89 & 0.8176 & 0.1447 \\
\gc DropGaussian+DualDiff3D & \gc \textbf{22.68} & \gc \textbf{0.7654} & \gc \textbf{0.1913} & \gc \textbf{24.21} & \gc \textbf{0.8154} & \gc \textbf{0.1606} & \gc \textbf{25.41} & \gc \textbf{0.8607} & \gc \textbf{0.1351} \\
\bottomrule
\end{tabular}
}
\end{table}

\section{Generalization Ability}
To verify DualDiff3D’s generalization ability, we conduct experiments on the LLFF dataset with 1/4 resolution and input views of 3, 6, and 9. We design three sets of comparative experiments to test its adaptability under different conditions, as follows:
\begin{itemize}
\item Standard 3DGS Baseline: We first reconstruct a 3DGS scene with 30k iterations, then enhance it with DIFIX3D and DualDiff3D for another 30k iterations each to test DualDiff3D’s enhancement effect on 3DGS.
\item Random Initialization: We skip 3DGS reconstruction, start from a random point cloud, and use DIFIX3D and DualDiff3D for 30k iterations to evaluate DualDiff3D’s adaptability to poor initialization.
\item Integration with SOTA Methods: We use DIFIX3D and DualDiff3D as plug-and-play modules on FSGS~\cite{fsgs} and DropGaussian~\cite{dropgaussian}. Both methods first reconstruct with 10k iterations, then enhance with another 10k iterations to test DualDiff3D’s adaptability to different base models.
\end{itemize}

\subsection{Impact of Initialization Quality on Generalization: Random vs. 3DGS}
As shown in \cref{tab:generalization}, DualDiff3D achieves strong performance even without reliable geometric initialization (\ie, random point cloud).
Notably, using 3DGS for initialization yields worse final metrics than random initialization.
This is because 3DGS is not designed for sparse-view settings, leading to severe misalignment and artifacts in its initial reconstruction, which severely constrain the subsequent optimization.
When DualDiff3D is initialized with 3DGS, the lower final metrics stem from the large “rectification cost” required to fix these inherent errors.
Most of the 30k enhancement iterations are spent correcting these errors instead of improving quality, even when the total number of iterations is doubled.

\subsection{Plug-and-Play Enhancement: Cross-Model Generalization}
When integrated into FSGS and DropGaussian, DualDiff3D stably and significantly improves reconstruction quality, while DIFIX3D is unstable and even degrades performance in some cases.
Diffusion-based refinement models inevitably introduce hallucinations and misalignment. 
DIFIX3D directly uses these flawed refined views without filtering, polluting the 3DGS optimization.
Our framework solves this by integrating the PSF strategy and CDW process as described in the DualDiff3D section. 
These modules ensure only reliable information is used, suppressing hallucinations and misalignment. This makes DualDiff3D a reliable plug-and-play module, confirming its good generalization ability.

\section{User Study}
We conducted a user study involving $30$ participants to subjectively evaluate the performance of 10 sets of reference views and novel views from our method and baseline approaches.
Specifically, the participants were tasked with assessing two key criteria: image quality and cross-view consistency. 
For each evaluation criterion, participants were asked to select the best image. 
We then calculated the proportion of each selected method.
The results of refinement and reconstruction tasks are presented in \cref{tab:user}, demonstrating that both DualDiff and DualDiff3D are more preferred by participants than the compared methods in terms of both image quality and cross-view consistency.

\vspace{-10pt}
\begin{table}[htbp]
\centering
\setlength\tabcolsep{5pt}
\renewcommand{\arraystretch}{1.0}
\captionsetup{type=table}
\caption{\textbf{User study of our method and baselines.}}
\label{tab:user}
\resizebox{0.8\linewidth}{!}{
\begin{tabular}{lcccc}
\toprule
\multirow{2}{*}{Method} & \multicolumn{2}{c}{Refinement} & \multicolumn{2}{c}{Reconstruction} \\
\cmidrule(lr){2-3} \cmidrule(lr){4-5}
& Quality & Consistency & Quality & Consistency \\
\midrule
3DGS       & 10.60\% & 21.00\% & 8.67\%  & 15.00\% \\
DIFIX3D    & 29.67\% & 32.33\% & 37.67\% & 33.00\% \\
\rowcolor[HTML]{F5F5F5}
DualDiff3D & 59.33\% & 46.67\% & 53.67\% & 52.00\% \\
\bottomrule
\end{tabular}}
\vspace{-30pt}
\end{table}

\section{Multi-view Consistency}
We further evaluate multi-view consistency of reconstructed novel views using the MEt3R~\cite{met3r} metric on LLFF~\cite{llff} and DL3DV~\cite{dl3dv}.
A lower MEt3R score indicates better geometric and appearance consistency across views.
As shown in \cref{tab:consistency}, DualDiff3D consistently outperforms 3DGS~\cite{3dgs} and DIFIX3D under all view-count settings on both datasets.

\vspace{-10pt}
\begin{table}[htbp]
\centering
\setlength\tabcolsep{5pt}
\renewcommand{\arraystretch}{1.0}
\captionsetup{type=table}
\caption{\textbf{Multi-view consistency (MEt3R$\downarrow$) comparisons.} Lower is better.}
\label{tab:consistency}
\resizebox{\linewidth}{!}{
\begin{tabular}{lccccccc}
\toprule
\multirow{2}{*}{\textbf{Method}} & \multicolumn{3}{c}{\textbf{LLFF}} & \multicolumn{4}{c}{\textbf{DL3DV}} \\
\cmidrule(lr){2-4} \cmidrule(lr){5-8}
& \textbf{3-View} & \textbf{6-View} & \textbf{9-View} & \textbf{3-View} & \textbf{6-View} & \textbf{9-View} & \textbf{24-View} \\
\midrule
3DGS    & 0.3212 & 0.2252 & 0.2109 & 0.2755 & 0.3224 & 0.3249 & 0.2834 \\
DIFIX3D & 0.2801 & 0.2025 & 0.1951 & 0.2652 & 0.3179 & 0.3183 & 0.2776 \\
\rowcolor[HTML]{F5F5F5}
DualDiff3D & \textbf{0.2635} & \textbf{0.1764} & \textbf{0.1615} & \textbf{0.2615} & \textbf{0.2720} & \textbf{0.2535} & \textbf{0.2133} \\
\bottomrule
\end{tabular}}
\vspace{-10pt}
\end{table}

\end{document}